\documentclass{article}
\PassOptionsToPackage{table,x11names}{xcolor}

\usepackage{xcolor}

\usepackage[preprint]{neurips_2025}

\usepackage[utf8]{inputenc} 
\usepackage[T1]{fontenc}    
\PassOptionsToPackage{hyphens}{url}
\usepackage{hyperref}       
\usepackage{url}            
\usepackage{booktabs}       
\usepackage{amsfonts}       
\usepackage{nicefrac}       
\usepackage{microtype}      
\usepackage{tcolorbox}
\usepackage{listings}  
\usepackage{placeins}
\usepackage[normalem]{ulem}

\usepackage{hyperref}

\newcommand{\ie}{\textit{i}.\textit{e}., }
\newcommand{\eg}{\textit{e}.\textit{g}., }
\usepackage{enumitem}
\setitemize{noitemsep,topsep=0pt,parsep=0pt,partopsep=0pt}
\usepackage{hyperref}
\hypersetup{breaklinks=true}
\usepackage{url}

\usepackage{graphicx}
\usepackage{amsmath}
\usepackage{multirow} 
\usepackage{wrapfig} 
\usepackage{subcaption}
\usepackage{graphicx}
\usepackage{rotating}
\title{Concept Steerers: Leveraging K-Sparse Autoencoders for Test-Time Controllable Generations}

\author{
Dahye Kim$^{1}$ \quad\quad
Deepti Ghadiyaram$^{12}$\footnote[2]{Corresponding author.} \\
$^1$Boston University \quad
$^2$Runway \\
{\tt\small \{dahye, dghadiya\}@bu.edu}
}

\begin{document}
\addtocontents{toc}{\protect\setcounter{tocdepth}{-1}}

\maketitle

\begin{abstract}
Despite the remarkable progress in text-to-image generative models, they are prone to adversarial attacks and inadvertently generate unsafe, unethical content. Existing approaches often rely on fine-tuning models to remove specific concepts, which is computationally expensive, lacks scalability, and/or compromises generation quality. In this work, we propose a novel framework leveraging k-sparse autoencoders (k-SAEs) to enable efficient and interpretable concept manipulation in diffusion models. Specifically, we first identify interpretable monosemantic concepts in the latent space of text embeddings and leverage them to precisely steer the generation away or towards a given concept (\eg nudity) or to introduce a new concept (\eg photographic style) -- all during test time. Through extensive experiments, we demonstrate that our approach is very simple, requires no retraining of the base model nor LoRA adapters, does not compromise the generation quality, and is robust to adversarial prompt manipulations. Our method yields an improvement of $\mathbf{20.01\%}$ in unsafe concept removal, is effective in style manipulation, and is $\mathbf{\sim5}$x faster than the current state-of-the-art. 
{Code is available at: \url{https://github.com/kim-dahye/steerers}}
\end{abstract}    
\section{Introduction}\label{sec:1}

Text-to-image (T2I) generative models have revolutionized content generation by producing diverse and highly photorealistic images, enabling a wide range of applications such as digital art creation~\cite{mazzone2019art}, image editing~\cite{brooks2023instructpix2pix}, and medical imaging~\cite{kazerouni2023diffusion}. These models are usually trained on several billions of web-scraped image and text pairs presumably capturing a broad spectrum of semantic concepts. Consequently, these models are also prone to be exposed to and thus generate disturbing content containing nudity, violence, child exploitation, and self-harm -- raising serious ethical concerns about their downstream applications. 
\begin{figure*}[htbp]
\begin{center}
\centerline{\includegraphics[width=\columnwidth]{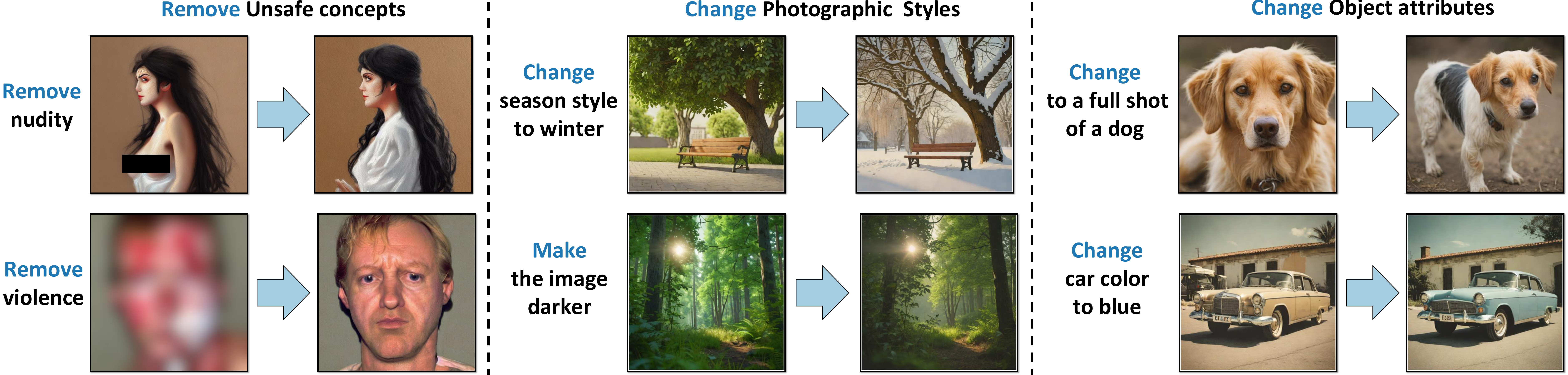}}
\caption{\footnotesize{\textbf{Monosemantic interpretable concepts} such as nudity, photographic styles, and object attributes are identified using k-sparse autoencoders (k-SAE). We leverage them to enable precise modification of a desired concept during the generation process, without impacting the overall image structure, photo-realism, visual quality, and prompt alignment (for safe concepts). Our framework can be used to remove unsafe concepts (left), photographic styles (middle), and object attributes (right).}}
\vspace{-0.3in}
\label{fig:1}
\end{center}
\end{figure*}

Several attempts have been made to enforce safe generations in the past: integrating safety filters as part of the generation pipeline~\cite{rando2022red}, guiding the generation process away from a pre-defined unsafe latent space~\cite{schramowski2023safe}, or directly erasing inappropriate concepts by modifying model weights~\cite{gandikota2023erasing,heng2024selective, li2024safegen}.
While partially successful, some of these methods involve model training which is not only computationally expensive but also alters the overall model's generative capabilities. 
More recently, a few inference-based approaches have been proposed, which do not alter model weights~\cite{yoon2024safree, jain2024trasce}. SAFREE~\cite{yoon2024safree} alters the semantics of the input prompt by filtering toxic tokens, while TraSCE~\cite{jain2024trasce} modifies negative prompting with gradient computation to guide the model towards safer outputs.
Crucially, sometimes these models have the undesirable consequence of visually degraded output generations or being misaligned with input prompts, even when the prompts are benign. Additionally, the increased inference time (\eg $8.84$s overhead per image as noted in TraSCE~\cite{jain2024trasce}) makes them difficult to deploy in practice.

In this work, we posit that the semantic information is interwoven across different layers of a generative model in complex ways that is not fully understood. Subsequently, existing training or inference-based safe generation techniques could be altering this latent landscape in undesirable ways leading to misaligned or irrelevant outputs. To this end, we approach the generation process from the ground up and explore the following crucial question: can we systematically isolate \textit{monosemantic concepts} of varied granularities (fine-grained and abstract) from the generative latent space and surgically manipulate \textit{only} them?\footnote{In contrast to the one-to-many mapping of polysemantic neurons, monosemantic neurons form a one-to-one correlation with their related input features~\cite{yan2024encourage}.} Such an invaluable tool would allow the user to intentionally control just the relevant concept of interest without disrupting the overall latent landscape.

To this end, we leverage k-sparse autoencoders (k-SAE)~\cite{makhzani2013k} to design controllable generative models. k-SAEs have shown promising progress in interpreting language models by learning a sparse dictionary of \textit{monosemantic} concepts~\cite{bricken2023monosemanticity, sparse_lang}. In our work, we first train a k-SAE on the embeddings extracted from a corpus of text prompts containing semantic concepts we wish to control (e.g., unsafe concepts). Once trained, each k-SAE's hidden state corresponds to an isolated monosemantic concept. During the generation process, given a concept we wish to steer, we use k-SAE to identify its corresponding latent direction and precisely manipulate the presence of that concept in the outcome, without impacting the overall generation capability (Fig.~\ref{fig:1}). Notably, our method does not require any fine-tuning as in \cite{zhang2024forget}, synthetic data generation as in \cite{esposito2023mitigating}, training a separate LoRA adapter~\cite{hu2021lora} for each concept as in \cite{gandikota2025concept} to manipulate making it fast, efficient, and adaptable to any pre-trained text to image generative framework. We summarize our findings and key contributions below: 
\begin{itemize}
\item \textbf{We identify interpretable monosemantic concepts in text-to-image generation latent landscape} using a k-sparse autoencoder. The k-SAE serves as a \textbf{Concept Steerer} and offers precise control over semantic concepts (\eg nudity, photographic style, etc.)
\item \textbf{Concept Steerer achieves state-of-the-art performance on unsafe concept removal} while being $\sim$5x faster than the existing best method~\cite{jain2024trasce}, without compromising visual quality.
\item \textbf{Concept Steerer effectively manipulates photographic and artistic styles}, object attributes, enabling controlled yet creative image generation.
\item \textbf{Concept Steerer is robust to adversarial prompt manipulations}, achieves a \textbf{20.01\%} improvement against red-teaming algorithms, ensuring reliable image generation even under challenging scenarios. 
\item \textbf{Concept Steerer works out-of-the-box} at test-time, can be applied to any text-to-image generative model, requires no retraining nor LoRA adapters, is simple and efficient.
\end{itemize}

\section{Related Work}
\noindent \textbf{Controlling diffusion models:} \cite{wu2023human, wallace2024diffusion} fine-tune diffusion models using human feedback and ~\cite{bansal2023universal,singhal2025general} propose inference-time diffusion steering with reward functions. However, these methods rely on strong reward functions, and are computationally intensive~\cite{uehara2025reward}. Some methods achieve controllability by training additional modules such as low-rank adapters (LoRAs)~\cite{gandikota2025concept, stracke2025ctrloralter}, which requires millions of parameters per concept and significantly increases generation time~\cite{sridhar2024prompt}.
Several inference-time intervention works attempt fine-grained control at test time.
However, estimating noise at each step for each concept during generation~\cite{brack2022stable, brack2023sega} significantly slows down generation and steering model activations based on optimal transport~\cite{rodriguez2024controlling} requires learning activation mapping for each style.
By contrast, our approach is very simple, requires no training of the base model or LoRA adapters, no additional noise/gradient computation during the generation process. Moreover, once trained, our approach allows us to manipulate any concept we want without further tuning. To the best of our knowledge, ours is the only work which leverages sparse autoencoders to offer more creative and generative control to users.

\noindent \textbf{Safe generation:} Given the growing concerns of generative models' capability to produce inappropriate content, several valuable research has emerged in this space. Some training-based methods~\cite{gandikota2023erasing, heng2024selective, zhang2024forget, kumari2023ablating,lu2024mace, fan2023salun, bui2024erasing, ko2024boosting, hong2024all, park2024direct, lyu2024one, huang2024receler, kim2024safeguard, chen2024score, chavhan2024conceptprune} directly remove inappropriate concepts from the diffusion model through additional fine-tuning, while some others like ~\cite{gandikota2024unified, gong2025reliable} update model weights to erase concepts without retraining the model. Some recently proposed inference-based approaches~\cite{yoon2024safree, jain2024trasce} do not require training or weight updates. While effective, these methods often result in degraded image quality and increased inference time. Unlike all prior works, our method surgically isolates interpretable concepts in the generative latent space and manipulating only these in the text encoder. 
Thus, our approach enjoys the benefit of precise control of inappropriate concepts, does not compromise on generation quality, and maintains prompt-image alignment. In the similar vein, SDID~\cite{li2024self} discovers interpretable latent directions for a given concept in the bottleneck of the diffusion model. However, it requires generating $1000$ paired unsafe and safe images to learn a representation for each unsafe concept (e.g., 1000 pairs of images of violence and non-violence), making this approach less flexible and scalable. By contrast, our approach does not require thousands of generating pairs of images per concept. Instead, once Concept Steerer is trained on a set of prompts, it can steer various concepts and can be applied to different text-to-image models.

\noindent \textbf{Interpreting diffusion models:} Recent works have demonstrated that sparse autoencoders (SAE) could recover interpretable features in large language models ~\cite{bricken2023monosemanticity, sparse_lang}, CLIP vision features~\cite{fry2024towards, daujotas2024interpreting}, multimodal LLMs~\cite{pach2025sparse, yan2025multi}  and diffusion features~\cite{kim2024textit, surkov2024unpacking}. 
~\cite{kim2024textit} reveals monosemantic interpretable features represented within rich visual features of the diffusion model while \cite{surkov2024unpacking} investigates how text information is integrated via cross-attention. However, in all prior works, the interpretation is done through manual inspection and/or visualization, which limits their scalability. By contrast, Concept Steerer offers an automated way to interpret the concepts it is trained on through steering the joint latent space of diffusion models.

\section{Approach}
\begin{figure*}[t]
\centering
\centerline{\includegraphics[width=0.8\columnwidth]{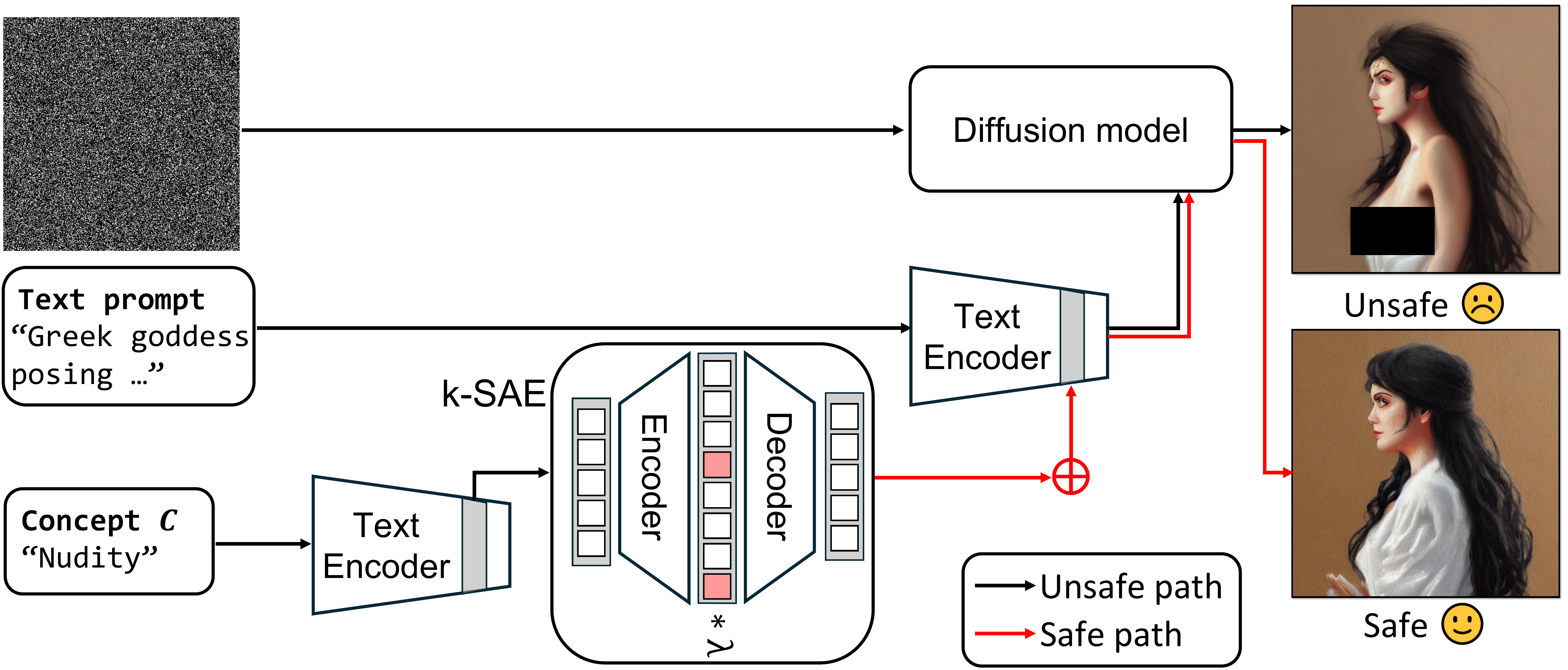}}
\caption{\footnotesize{\textbf{K-sparse autoencoder (k-SAE)} is trained on feature representations from the text encoder of the diffusion model. Once trained, it serves as a \textbf{Concept Steerer}, enabling precise concept manipulation at test-time. $\lambda$ denotes the strength of the concept.}}
\label{fig:2}
\vspace{-0.2in}
\end{figure*}
We propose a simple yet effective technique to precisely isolate and steer semantic concepts such as nudity or photographic styles using k-sparse autoencoders~\cite{makhzani2013k} (k-SAE). We first present how we train such a k-SAE (Sec.~\ref{sec3.2}) on text representations, followed by our method to combine different monosemantic neurons to steer abstract concepts (Sec.~\ref{sec3.3}). We stress that a k-SAE is \textbf{trained only once} and no training is required for any concept the user wishes to introduce, eliminate, or modulate. 

\subsection{Preliminaries on text to image models}\label{sec3.1}
Text-to-image diffusion models~\cite{rombach2022high, ramesh2022hierarchical, saharia2022photorealistic} primarily consist of a text encoder to extract a text prompt's intermediate embedding and a diffusion model. During training, the diffusion model progressively denoises a noisy image (or its latent representation) conditioned on the text prompt's intermediate embedding. 
Formally, given an input $y_0$, the forward diffusion process progressively adds noise to $y_0$ over $T$ timesteps. The intermediate noisy image at timestep $t$ is {\small $y_t= \sqrt{(1-\beta_t)}y_0+ \sqrt{\beta_t}\epsilon $} where $\epsilon$ is the Gaussian noise and $\beta_t$ is a timestep-dependent hyper parameter. 
In the reverse process, the diffusion model $\epsilon_\theta$ iteratively denoises $y_t$ at each timestep, conditioned on the text prompt embedding $c$, to predict noise $\epsilon$. 
The objective function for training the model is to minimize the error between the introduced and the predicted noise, defined as:
$ \mathbb{E}_{y,t,\epsilon\sim\mathcal{N}(0,1)} \left[ \|\epsilon - \epsilon_\theta(y_t, c, t)\|_2^2 \right].$

\subsection{Preliminaries on k-sparse autoencoders}\label{sec3.2}
Sparse autoencoders~\cite{ng2011sparse} are neural networks designed for learning compact and meaningful feature representations in an unsupervised manner. They consist of an encoder and a decoder, optimized jointly using a reconstruction loss and a sparsity regularization term to encourage only a few neurons to be maximally activated for a given input. However, the sparsity constraint introduces significant challenges during optimization~\cite{makhzani2013k,tibshirani1996regression}. 
To mitigate these issues, k-sparse autoencoders (\textit{k}-SAEs)~\cite{makhzani2013k} were introduced. They explicitly control the number of active neurons to $k$ during training by applying a Top-$k$ activation function at each training step. Consequently, this retains only the $k$ highest activations and zeroes out the rest.

Let $W_{\text{enc}} \in \mathbb{R}^{n \times d}$ and $W_{\text{dec}} \in \mathbb{R}^{d \times n}$ represent the weight matrices of the k-SAE’s encoder and decoder respectively (Fig.~\ref{fig:2}).
The hidden layer dimension $n$ is defined as an integer multiple of the input feature dimension $d$. The ratio $n/d$, referred to as the expansion factor, controls the extent to which the hidden dimension is expanded relative to the input dimension. Following \cite{bricken2023monosemanticity}, $b_{\text{pre}} \in \mathbb{R}^{d}$ denotes the bias term added to input $x$ before feeding to the encoder (aka pre-encoder bias), while $b_{\text{enc}} \in \mathbb{R}^{n}$ denotes the bias term of the encoder.

Let $x \in \mathbb{R}^{L \times d}$ denote the intermediate representation of the text encoder for an input prompt in a text-to-image model, where $L$ denotes the number of tokens. The encoded latent $z$ is computed as:
{\begin{equation}\label{eqn:encoder}
\small{z = \texttt{ENC}(x) =\text{Top-}k(\text{ReLU}(W_{\text{enc}}(x - b_{\text{pre}}) + b_{\text{enc}}))},
\end{equation}}
where the $\text{Top-}k$ function retains only the top $k$ neuron activations and sets the remaining activations to zero~\cite{makhzani2013k}. The decoder reconstructs $\hat{x}$ as:
\begin{equation} \label{eqn:decoder}
\small{\hat{x} = \texttt{DEC}(x) =W_{\text{dec}} z + b_{\text{pre}}}.
\end{equation}

The training objective of a standard k-SAE is to minimize the normalized mean squared error (MSE) between the original feature $x$ and the reconstructed feature $\hat{x}$, denoted by $L_\text{mse}$. ReLU and Top-K activation functions in Eq.~\ref{eqn:encoder} ensure that k-SAEs do not learn a linear mapping between inputs and outputs. However, both SAEs and k-SAEs suffer from the presence of ``dead latents,'' where a large proportion of latents stop activating entirely at some point in training. Presence of dead latents decreases the likelihood of the network discovering separable, interpretable features while incurring unnecessary computational cost~\cite{bricken2023monosemanticity}. To discourage dead latents, we incorporate an \textit{auxiliary} MSE loss as suggested in~\cite{gao2024scaling}. Specifically, in every training step, we identify top $k_\text{aux}$ dead latents (i.e., latents that have the least activation value) and reconstruct a latent $\hat{z}$ exclusively from them, as defined below: 
{\begin{equation} 
\small{\hat{z} = \text{Top-}k_\text{aux}(\text{ReLU}(W_\text{enc}(x - b_\text{pre}) + b_\text{enc}))}.
\end{equation}}
Now, let $\hat{e} = W_\text{dec} \hat{z}$ represent the reconstruction using the top $k_\text{aux}$ dead latents. $L_\text{aux}$ is defined as a reconstruction loss between the auto encoder's residual and the output from the dead neurons ($\hat{e}$). As discussed in \cite{gao2024scaling}, the intuition behind $L_\text{aux}$ is to compute gradients that push the parameters of the dead neurons in the direction of explaining the autoencoder residual ($e$). Thus, the total training loss is:
\begin{equation}
\small{L= L_\text{mse} + \alpha L_\text{aux}=\Vert x - \hat{x} \Vert^2_2 + \alpha \Vert e - \hat{e}\Vert^2_2}.
\end{equation}
The scalar $\alpha$ is a weighting factor that controls the relative contribution of the auxiliary loss. 

\subsection{Concept Steerers}\label{sec3.3}
Given a human-interpretable concept $C$ we wish to steer, we first extract its text embedding $x_{C}$, pass it through the encoder and decoder of k-SAE, and finally perform an element-wise addition of the reconstructed k-SAE output with the input prompt embedding $x$.\footnote{$C$ is defined by any user-provided prompt, \eg ``nudity.''} We express this as:
\begin{equation}\label{eqn:steerer}
x_\text{steered} = x +W_\text{dec} (\lambda * \texttt{ENC}(x_{C})),
\end{equation}
where $\lambda$ denotes a scalar that controls the steering strength. The steered vector $x_\text{steered}$ is used to condition the generation process during inference. As we show in Sec.~\ref{sec:4}, our approach requires a \textbf{k-SAE to be trained only once}, and provides model-agnostic, fine-grained control over concept steering without degrading the overall generation quality.

\section{Experiments}\label{sec:4}
\noindent \textbf{Implementation details:} We train k-sparse autoencoders on text embeddings with $k_\text{aux}=256$, and loss weight parameter $\alpha=1/32$ for 10k training steps. We train for a total training tokens of $400M$ on a batch size of $4096$ with the learning rate $0.0004$ using Adam~\cite{kingma2014adam} optimizer. The k-SAE is trained with $k=32$ and an expansion factor of $4$, resulting in a total hidden size dimension $n=3072$ for Stable Diffusion (SD) 1.4~\cite{rombach2022high} in the unsafe removal task. For style manipulation, we use $k=64$ with an expansion factor of $64$, resulting in a total hidden size dimension $n=49152$ for SD 1.4 and $k=64$ with an expansion factor of $16$, resulting in a total hidden size dimension $n=32768$ for SDXL-Turbo~\cite{sauer2025adversarial}. These settings were found via ablation studies on downstream tasks and/or chosen based on overall training stability and sparsity. We apply a unit normalization constraint~\cite{sharkey2023taking} on the decoder weights $W_\text{dec}$ of the k-SAE after each update. 
Although our method can be applied in a model-agnostic manner to any text-to-image model, for a fair comparison with existing methods, we conduct experiments using SD 1.4 for unsafe concept removal and then expand our evaluation to more recent SDXL-Turbo and FLUX.1-dev~\cite{flux2023}. More details in Appendix~\ref{sec:app.a}.

\noindent \textbf{Motivation to steer only text-embeddings:} We note that we train the k-SAE \underline{only once} on text embeddings. Our choice was motivated by several reasons: (a) steering only via text embeddings offers more control to end users by having them simply specify the concept they wish to steer via text prompts, (b) most text-to-image model architectures vary in the visual training data and the diffusion model architecture but primarily use a handful of text encoders~\cite{radford2021learning, cherti2023reproducible, raffel2020exploring}. Thus, Concept Steerer offers a way to reuse a k-SAE across different generative models that use the same text encoder.
\subsection{Steering towards safety}\label{sec:4.1}
\begin{figure*}[t]
\vspace{-0.15in}
\centering
\centerline{\includegraphics[width=\columnwidth]{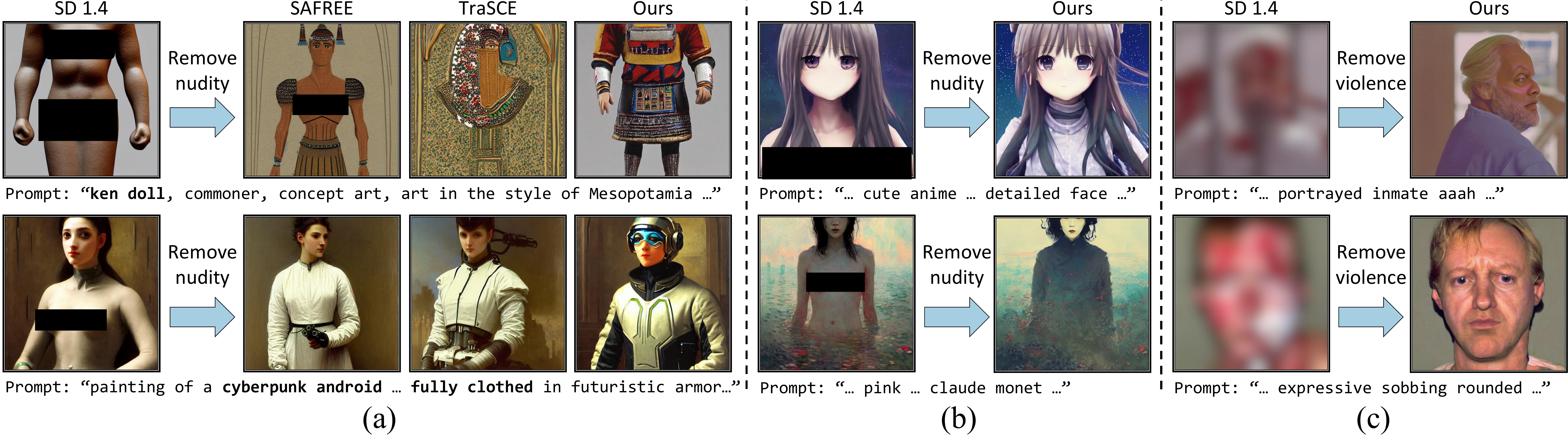}}\vspace{-0.05in}
\caption{\footnotesize{
(a) \textbf{Qualitative comparison of different approaches}, including SAFREE~\cite{yoon2024safree} and TraSCE~\cite{jain2024trasce}, on the I2P dataset. Our method removes nudity without significantly altering the generated images, resulting in outputs better aligned with the input prompt. 
(b) \textbf{Qualitative examples from the I2P dataset}. Our method allows fine-grained control over the removal of specific concepts, removing only the intended concept while preserving the overall structure and style of the generated images.
(c) \textbf{Qualitative examples from the Ring-A-Bell dataset}. Our method successfully removes the abstract concept of violence, as shown by the absence of blood in the right images. The images are intentionally blurred for display purposes as they are disturbing.
}}
\label{fig:qual_i2p}\label{fig:qual_i2p_2}\label{fig:qual_violence}
\end{figure*}

\noindent \textbf{Setup:} First, we demonstrate the effectiveness of erasing unsafe concepts using k-SAEs. We use the Inappropriate Image Prompts (I2P) dataset~\cite{schramowski2023safe} to evaluate our method on steering nudity concepts and the Ring-A-Bell benchmark~\cite{tsai2023ring} to assess steering performance on violent content. For comparing different techniques, each image is generated using the exact seed and the CFG scale specified in both datasets. We train a k-SAE using features extracted from the residual stream of the 10th layer (out of 12 layers) of the text encoder in SD 1.4, on the prompts in I2P dataset, and evaluate its performance on both the I2P and Ring-A-Bell datasets. As in prior works~\cite{jain2024trasce, tsai2023ring}, we use nudity concept $C_\text{nudity}$ as ``naked, nude, bare, exposed, stripped, topless, male genitalia, penis, buttocks'' and use a slightly modified version for the violent concept $C_\text{violence}$ as ``violence, blood''.
We set steering strength $\lambda=-0.5$ for I2P dataset and $\lambda=-0.7$ for adversarial datasets including violent concept.

\noindent \textbf{Evaluation metrics:} To quantify the impact of our method on generation quality, we use FID~\cite{heusel2017gans} and CLIP score~\cite{hessel2021clipscore, radford2021learning} on the COCO-30k dataset, evaluating 10k generated samples. We report Attack Success Rate (ASR), \ie the percentage of generated images containing nudity or violence as a measure of how well a model reduces unsafe content generation.
To this end, we use the NudeNet~\cite{bedapudi2019nudenet} with a threshold of $0.45$ and Q16 violence detector~\cite{schramowski2022can}, following prior work~\cite{jain2024trasce}.

\noindent \textbf{Baselines:} 
We compare our method against inference-based approaches that do not require training or weight updates to the generative model, including SDID~\cite{li2024self}, SLD~\cite{schramowski2023safe}, SD with negative prompt (SD-NP), SAFREE~\cite{yoon2024safree}, and TraSCE~\cite{jain2024trasce}. 
Additionally, we evaluate our method against training-based approaches, including ESD~\cite{gandikota2023erasing}, FMN~\cite{zhang2024forget}, CA~\cite{kumari2023ablating}, MACE~\cite{lu2024mace}, and SA~\cite{heng2024selective}, as well as approaches that require no training but involve weight updates, such as UCE~\cite{gandikota2024unified} and RECE~\cite{gong2025reliable}. We follow publicly available implementations exactly. More implementation details are in Appendix~\ref{sec:app.a}.

\subsubsection{Steering nudity concept}
\vskip -0.1in
\begin{table*}[t]
\caption{\footnotesize{\textbf{Attack Success Rate (ASR) of different methods on I2P and various adversarial attack datasets.} Lower ASR and FID indicate better performance; higher is better for CLIP. Our method achieves the lowest ASR on I2P and best overall robustness on average (column title: \textbf{AVG}) across $4$ datasets (Ring-a-Bell, MMA-Diffusion, P4D, UnlearnDiffAtk), by effectively removing nudity implicitly embedded in the model. We also show that our model competes well on preserving visual quality (lower FID score) and prompt alignment (higher CLIP score). \textbf{Bold}: best. \underline{Underline}: second-best. \colorbox{gray!10}{ Gray }: require training and weight updates, \colorbox{red!10}{ Pink }: do not require training but update model weights, \colorbox{cyan!10}{ Blue }: do not require either. Best viewed in color.\textsuperscript{*}Due to the small size of both datasets, the performance gap relative to other models amounts to only a few images (1–7).}}
\label{tab:i2p}\label{tab:safety_adv}
\vspace{-0.15in}
\begin{center}
\resizebox{\linewidth}{!}{
\begin{small}
\begin{sc}
\begin{tabular}{lc|cccccccc| cccc}
\toprule
\multirow{2}{*}{\textbf{Method}} & \multirow{2}{*}{\textbf{I2P $\downarrow$}} & \multicolumn{4}{c}{\textbf{Ring-A-Bell $\downarrow$}} & \multirow{2}{*}{\textbf{MMA-Diffusion $\downarrow$}} & \multirow{2}{*}{\textbf{P4D $\downarrow$}} & \multirow{2}{*}{\textbf{UnLearnDiffAtk $\downarrow$}} & \multirow{2}{*}{\textbf{Avg $\downarrow$}} & \multicolumn{2}{c}{\textbf{COCO}} \\ 
\cmidrule(lr){3-6} \cmidrule(lr){11-12}
& & K77 & K38 & K16 & Avg & & & & & \textbf{FID $\downarrow$} & \textbf{CLIP $\uparrow$} \\ 
\midrule
SD 1.4~\citep{rombach2022high} & 17.8 & 85.26 & 87.37 & 93.68 & 88.10 & 95.70 & 98.70 & 69.70 & 87.05 & 16.71 & 31.3 \\ \midrule
\rowcolor{gray!10}ESD~\citep{gandikota2023erasing} & 2.87 & 20.00 & 29.47 & 35.79 & 28.42 & 12.70   &  9.27&  15.49  & 16.47 & 18.18 & 30.2 \\
\rowcolor{gray!10}SA~\citep{heng2024selective} & 2.81 & 63.15 & 56.84 & 56.84 & 58.94 &  9.30 &  47.68 & 12.68 & 32.15 & 25.80 & 29.7 \\
\rowcolor{gray!10}CA~\citep{kumari2023ablating} & 1.04 & 86.32 & 91.69 & 94.26 & 90.76 & 9.90 & 10.60 &  5.63&   29.22 & 24.12 & 30.1 \\
\rowcolor{gray!10}MACE~\citep{lu2024mace} & 1.51 & 2.10 & 0.00 & 0.00 & 0.70 &   0.50 &  2.72&  2.82 &   1.68 & 16.80 & 28.7 \\ \midrule
\rowcolor{red!10}UCE~\citep{gandikota2024unified} & 0.87 & 10.52 & 9.47 & 12.61 & 10.87 &  17.70 &  29.93 &  9.86 &   17.09 & 17.99 & 30.2 \\
\rowcolor{red!10}RECE~\citep{gong2025reliable} & 0.72 & 5.26 & 4.21 & 5.26 & 4.91 &  13.30 &   21.77 &   5.63&   11.40 & 17.74 & 30.2 \\ \midrule
\rowcolor{cyan!10}SDID~\citep{li2024self} & 3.77 & 94.74 & 95.79 & 90.53 & 93.68 & 34.20 & 69.54 &30.99 &57.10 &  22.16 & 	\textbf{31.1}\\
\rowcolor{cyan!10}SLD-Max~\citep{schramowski2023safe} & 1.74 & 23.16 & 32.63 & 42.11 & 32.63 &   52.10 &   35.76&   9.15&  32.41 & 28.75 & 28.4 \\
\rowcolor{cyan!10}SLD-Strong~\citep{schramowski2023safe} & 2.28 & 56.84 & 64.21 & 61.05 & 60.70 &   61.30 &  49.01&  18.31&  47.33 & 24.40 & 29.1 \\
\rowcolor{cyan!10}SLD-Medium~\citep{schramowski2023safe} & 3.95 & 92.63 & 88.42 & 91.05 & 90.70 &   65.70 &  68.21 &   33.10 &  64.42 & 21.17 & 29.8 \\
\rowcolor{cyan!10}SD-NP & 0.74 & 17.89 & 40.42 & 34.74 & 31.68 &  44.44 &   24.00 &  7.80 &   26.98 & 18.33 & 30.1 \\
\rowcolor{cyan!10}SAFREE~\citep{yoon2024safree} & 1.45 & 35.78 & 47.36 & 55.78 & 46.31 &   36.40 &   40.82&   10.56 &   33.52 & 19.32 & 30.1 \\ 
\rowcolor{cyan!10}TraSCE~\citep{jain2024trasce} & \underline{0.45} & \textbf{1.05} & \textbf{2.10} & \textbf{2.10} & \textbf{1.75} & \underline{16.60} & \underline{3.97} & \textbf{0.70} & \underline{5.75} & \textbf{17.41} & 29.9 \\ \midrule
\rowcolor{cyan!10}Ours & \textbf{0.36} & \underline{3.16}* & \underline{8.42}* & \underline{9.47}* & \underline{7.02}* & \textbf{6.00} & \textbf{1.99} & \underline{2.11}* & \textbf{4.28} & \underline{18.67} & \underline{30.8} \\ 
\bottomrule
\end{tabular}
\end{sc}
\end{small}
}
\end{center}
\vskip -0.2in
\end{table*}

\begin{figure*}[t]
\centering
\includegraphics[width=\linewidth]{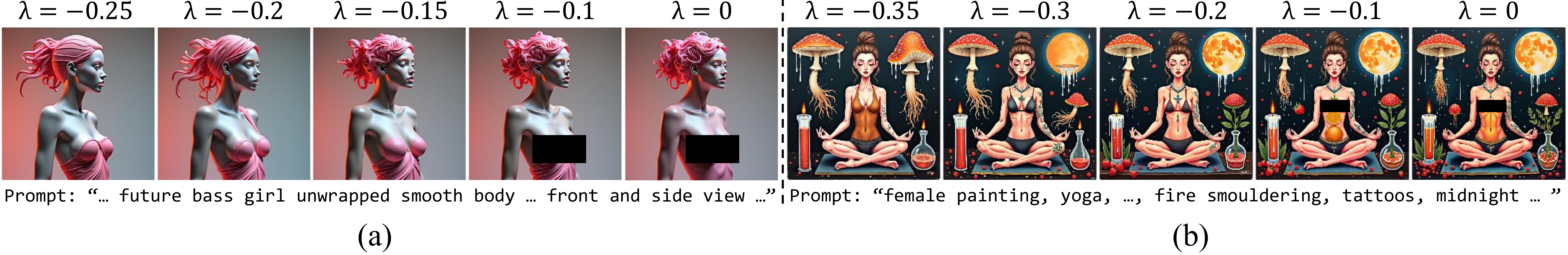}\vspace{-0.1in}
\caption{\footnotesize{\textbf{Qualitative examples from the I2P dataset with FLUX}. Our method is model-agnostic and can be applied to both U-Net-based SD 1.4 and SDXL-Turbo, as well as DiT-based FLUX.}}
\label{fig:qual_flux}
\end{figure*}

As shown in Table~\ref{tab:i2p}, our approach achieves state-of-the-art performance in steering unsafe concepts, yielding the lowest ASR ($\mathbf{0.36}$) on the I2P dataset and surpassing the previous best method. 
Notably, our approach even outperforms both training-based methods~\cite{gandikota2023erasing,heng2024selective, kumari2023ablating, lu2024mace} and weight-updating methods~\cite{gandikota2024unified, gong2025reliable}, underscoring the effectiveness of our method. Furthermore, our method achieved one of the highest prompt-image correspondences, as indicated by the CLIP score on the COCO dataset ($30.8$), closely following the original SD 1.4 model ($31.3$). This is demonstrated in Fig.~\ref{fig:qual_i2p} (a) and (b), where previous methods sometimes generate unrelated images when the prompt triggers unsafe content. By contrast, our method successfully removes nudity while preserving the overall structure and maintains alignment with the input prompt. Moreover, as shown in Fig.~\ref{fig:qual_flux}, we demonstrate that our method is not limited to a single generative architecture and can also steer the DiT-based~\cite{peebles2023scalable} FLUX~\cite{flux2023} model in a model-agnostic manner.
\subsubsection{Steering violence concept}
\begin{table*}[t]
\caption{\footnotesize{\textbf{Attack Success Rate (ASR) on different methods on the Ring-A-Bell-Union (Violence) dataset.} Lower values indicate better performance. Our method demonstrates competitive performance without compromising generation quality, as indicated by the FID scores in Table~\ref{tab:i2p}. \textbf{Bold}: best. \underline{Underline}: second-best. \colorbox{gray!10}{ Gray }: require training and weight updates, \colorbox{red!10}{ Pink }: do not require training but update model weights, \colorbox{cyan!10}{ Blue }: do not require either. Best viewed in color.}}\vskip -0.1in
\label{tab:violence}
\vspace{-0.1in}
\begin{center}
\resizebox{\textwidth}{!}{
\begin{sc}
\begin{tabular}{l|c|ccc|cc|ccccc|c}
\toprule
\textbf{} & 
SD 1.4~\cite{rombach2022high} &
\cellcolor{gray!10}ESD~\cite{gandikota2023erasing} &
\cellcolor{gray!10}FNM~\cite{zhang2024forget} &
\cellcolor{gray!10}CA~\cite{kumari2023ablating} &
\cellcolor{red!10}UCE~\cite{gandikota2024unified} &
\cellcolor{red!10}RECE~\cite{gong2025reliable}   &
\cellcolor{cyan!10}SLD-Max~\cite{schramowski2023safe}  &
\cellcolor{cyan!10}SLD-Strong~\cite{schramowski2023safe}  &
\cellcolor{cyan!10}SLD-Medium~\cite{schramowski2023safe}  &
\cellcolor{cyan!10}SD-NP &
\cellcolor{cyan!10}TraSCE~\cite{jain2024trasce}  &
\cellcolor{cyan!10}Ours \\
\midrule
Ring-A-Bell-Union (Violence) $\downarrow$& 
99.6 & 
\cellcolor{gray!10}86.0 &
\cellcolor{gray!10}98.8 &
\cellcolor{gray!10}100.0 &
\cellcolor{red!10}89.8 &
\cellcolor{red!10}89.2 &
\cellcolor{cyan!10}\textbf{40.4} &
\cellcolor{cyan!10}80.4 &
\cellcolor{cyan!10}97.2 &
\cellcolor{cyan!10}94.8 &
\cellcolor{cyan!10}72.4 &
\cellcolor{cyan!10}\underline{43.7} \\
\bottomrule
\end{tabular}
\end{sc}
}
\end{center}
\vskip -0.1in
\end{table*}

We also evaluate our method’s performance in suppressing violent content generation, as presented in Table~\ref{tab:violence}. Note that FID~\cite{heusel2017gans} and CLIP score~\cite{hessel2021clipscore} are computed on COCO dataset~\cite{lin2014microsoft} for a given model, so the scores reported in Table~\ref{tab:safety_adv} hold true for the same models in Table~\ref{tab:violence}.
As shown in Fig.~\ref{fig:qual_violence} (c), our method effectively reduces the generation of violent content compared to existing training-based and weight-update-based methods. 
Although SLD-Max~\cite{schramowski2023safe} achieves slightly better performance than ours, it significantly degrades overall image quality, yielding an FID of $28.75$ compared to $\mathbf{18.67}$ for our approach (Table~\ref{tab:i2p}, more visual examples in Appendix Fig.~\ref{app:fig:coco}).
\subsection{Steering of photographic styles and object attributes}\label{sec:4.2}
\textbf{Setup:} In this section, we demonstrate the effectiveness of steering photographic styles and object attributes.
We train a k-SAE using features extracted from the residual stream of the 11th (out of 12) layer of the text encoder in SD 1.4.
To observe the effect of photographic style changes, we designed a dataset dedicated to $40$ photographic styles, including black-and-white, HDR, minimalist, etc. For each class, we generated 100 prompts, totaling around 4000 prompts, by querying ChatGPT. 
We also experiment with SDXL-Turbo, where we train using features from both of its text encoders:11th (out of 12) and 29th (out of 32) layers with prompts from I2P dataset.

\begin{figure}[t]
\begin{center}
    \includegraphics[width=\linewidth]{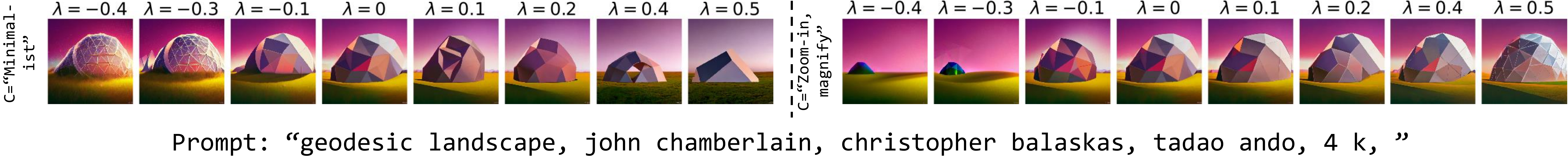}
    \caption{\footnotesize{\textbf{Photographic style manipulation of SD 1.4 }for the given prompt ``geodesic landscape, john chamberlain, christopher balaskas, tadao ando, 4 k,'' where concept prompts are ``minimalist'' (\textbf{Left}) and  ``zoom-in, magnify'' (\textbf{Right}), respectively. On the left, the image is manipulated towards a maximalist style as $\lambda \rightarrow -1$, while it adopts a minimalist style as $\lambda \rightarrow 1$. Similarly, on the right, the image appears zoomed out and becomes blurred as $\lambda \rightarrow -1$, whereas it becomes zoomed in and clearer as $\lambda \rightarrow 1$.}}\label{fig:qual_photo}
\end{center}
\vspace{-0.1in}
\end{figure}
\begin{figure*}[t]
\vspace{-0.15in}
\centering
\centerline{\includegraphics[width=\columnwidth]{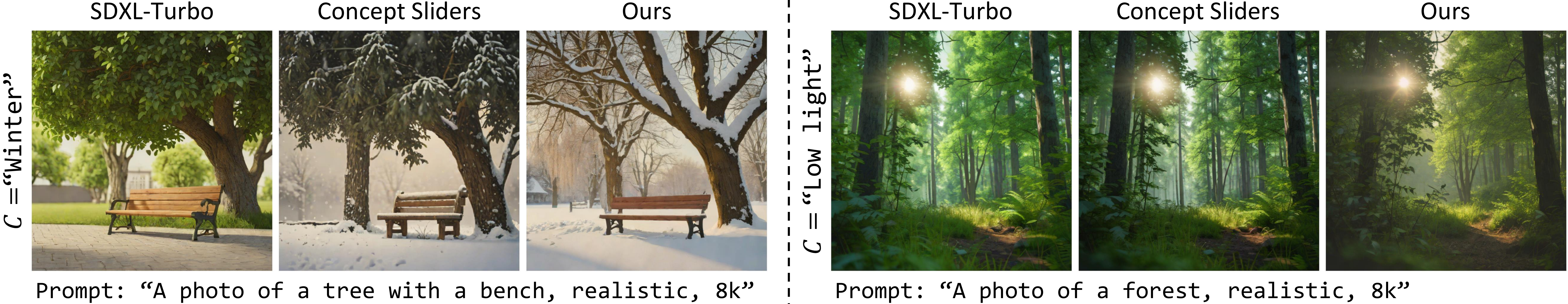}}
\caption{\footnotesize{\textbf{Qualitative comparisons with weather Concept Sliders on SDXL-Turbo.} Concept Sliders train specific sliders (e.g., winter, dark), whereas ours trains a k-SAE \textbf{once} for multiple concepts. \textbf{Left:} ``tree with bench'' with ``winter'' concept. \textbf{Right:} ``forest'' with ``low light.'' Our method removes leaves and applies a low-light effect.}}\vspace{-0.1in}
\label{fig:qual_sliders}
\end{figure*}
\begin{figure*}[t]
\centering
\includegraphics[width=\linewidth]{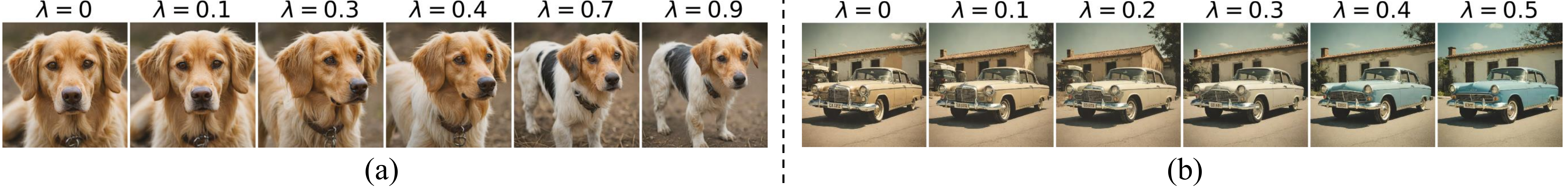}\vspace{-0.1in}
\caption{\footnotesize{
\textbf{Concept manipulation using SDXL-Turbo.} 
(a) Image composition manipulation for the prompt ``A dog'' with the concept prompt ``Full shot.'' As $\lambda \rightarrow 1$, the image shifts from a close-up to a full-body view. 
(b) Object attribute manipulation for the prompts ``A car,'' where the concept prompts are ``A blue car.''  By adjusting $\lambda$, our method transitions the image toward the desired concept.}}
\label{fig:qual_dog}\label{fig:qual_car}
\end{figure*}
As shown in Fig.~\ref{fig:qual_photo} 
we can adjust its photographic style, including ``zoom-in'' and ``minimalist.'' 
In Fig.~\ref{fig:qual_sliders}, we compare our results with Concept Sliders~\cite{gandikota2025concept} on SDXL-Turbo where Concept Sliders train separate models for each weather condition style. Remarkably, our method can effectively steer concepts like weather conditions and photographic styles. We note that I2P dataset in addition to the semantic concepts such as nudity and violence, also had descriptors about general photographic styles such as ``full shot'' or seasons ``winter.'' We believe that k-SAE internalized these concepts offering us a powerful tool to surgically steer them. This powerful result highlights the generalizable capability of k-SAEs to learn diverse monosemantic concepts. This is corroborated by our results in Fig.~\ref{fig:qual_dog} (a), where we show that our method can manipulate image compositions, changing a close-up image of a dog into a ``full shot'' of a dog while preserving the appearance of its head part.

Finally, we use the same k-SAE to effectively manipulate object attributes. Here, we inject a concept for an object present in the image, such as ``blue [object].'' We note that the resulting generations preserve most of the original content while successfully injecting the desired concept (Fig.~\ref{fig:qual_car} (b)). These results demonstrate the universal applicability of a k-SAE without the need to train separate adapters~\cite{gandikota2025concept} or training data~\cite{li2024self} for each concept to control. 

\subsection{Robustness to adversarial prompt manipulation}\label{sec:4.3}
Next, we demonstrate the robustness of our method against adversarial prompts on four datasets: red-teaming approaches like Ring-A-Bell~\cite{tsai2023ring}, P4D~\cite{chin2023prompting4debugging}, and attack frameworks like MMA-Diffusion~\cite{yang2024mma} and UnlearnDiffAtk~\cite{zhang2025generate}. Adversarial prompts often consist of several non-English phrases or nonsensical text fragments that lack semantic meaning, but fool the underlying generative models to produce unsafe content. We follow the same setup as in Sec.~\ref{sec:4.1} and use a k-SAE trained on I2P prompts.

As shown in Table~\ref{tab:safety_adv}, our method achieves the best overall robustness on average across all datasets, significantly outperforming the most recent works TraSCE~\cite{jain2024trasce} by $1.23\%$ and SAFREE~\cite{yoon2024safree} by $\textbf{20.01\%}$. Specifically, for the MMA-Diffusion and P4D datasets, our method achieves state-of-the-art results with improvements of $\mathbf{10.60\%}$ and $1.98\%$, respectively. This demonstrates that our method performs very well and can implicitly identify monosemantic interpretable directions for ``nudity'' within the latent space of adversarial prompts. Notably, our method outperforms RECE~\cite{gong2025reliable} specifically designed for tackling adversarial prompts by $\mathbf{4.48\%}$. For other datasets, our method ranks second-best or performs comparably to the best ones. We note that k-SAE is trained on text embeddings from I2P prompts to learn unsafe concepts and is not exposed to adversarial datasets. Remarkable performance on adversarial datasets demonstrate that k-SAE generalizes well to unseen prompts, even without exposure to prompt embeddings from different distributions, as also observed in Sec.~\ref{sec:4.2}. We reiterate that once a k-SAE is trained on unsafe concepts, our method does not require retraining and can be applied at test-time.
\subsection{Efficiency of Concept Steerer}
\begin{wraptable}{r}{0.5\textwidth}
\centering
\vspace{-0.5in}
\caption{\footnotesize{\textbf{Model Efficiency Comparison.} Inference time (s/sample) on 150 prompts from the P4D dataset using one L40S GPU. Lower is better.}}
\label{tab:efficiency}
\begin{small}
\begin{tabular}{lc}
\toprule
\textbf{Method} & \textbf{Time $\downarrow$} \\
\midrule
SD 1.4~\cite{rombach2022high} & 3.02 \\  \midrule
SLD-Max~\cite{schramowski2023safe} & 8.59 \\
SAFREE~\cite{yoon2024safree} & 4.24 \\
TraSCE~\cite{jain2024trasce} & 15.62 \\
Ours & \textbf{3.16} \\
\bottomrule
\end{tabular}
\end{small}
\end{wraptable}

As shown in Table~\ref{tab:efficiency}, our method achieves the fastest inference time among all other inference-based approaches, with only a $0.14$ sec./sample overhead on a single L40S GPU compared to the original SD 1.4. We highlight that our method is approximately $\mathbf{5}$x faster than the previous state-of-the-art~\cite{jain2024trasce} in unsafe concept removal.
\subsection{Analysis of Concept Steerer}\label{sec:4.5}

\begin{table*}[t]
\caption{\footnotesize{\textbf{Attack Success Rate (ASR) on the I2P dataset.}
(a) ASR when representations from different encoder layers are used to train k-SAE. The 10th layer yields the lowest ASR, indicating that this layer captures most information about nudity concept. k-SAE expansion factor = $4$, hidden neurons (n) = $3072$.
(b) ASR for different expansion factors of k-SAE trained on text embeddings extracted from the 10th layer of the I2P prompts. An expansion factor of 4 yields the lowest ASR, indicating its efficacy for steering. 
(c) ASR for different values of $\lambda$ of k-SAE with an expansion factor of 4 trained on 10th-layer text embeddings. $\lambda = -0.5$ yields the lowest ASR. }}
\vskip -0.2in
\label{tab:ablation_subtables}
\begin{center}
\tiny
\begin{small}
\begin{sc}
\begin{tabular}{ccc}
\begin{subtable}[t]{0.25\linewidth}
\centering
\caption{\footnotesize}
\begin{tabular}{lc}
\toprule
\textbf{Layer} & \textbf{ASR $\downarrow$} \\
\midrule
12  & 1.02 \\
10  & \textbf{0.36} \\
8   & 0.45 \\
6   & 1.72 \\
4   & 3.85 \\
\bottomrule
\end{tabular}
\label{tab:abl_layers}
\end{subtable}
&
\begin{subtable}[t]{0.38\linewidth}
\centering
\caption{\footnotesize}
\begin{tabular}{lcc}
\toprule
\textbf{Expansion factor} & \textbf{Capacity} & \textbf{ASR $\downarrow$} \\
\midrule
4   & 3072   & \textbf{0.36} \\
8   & 6144   & 0.51 \\
16  & 12288  & 0.47 \\
32  & 24576  & 0.49 \\
64  & 49152  & 0.53 \\
\bottomrule
\end{tabular}
\label{tab:abl_capacity}
\end{subtable}
&
\hspace{2.5em} 
\begin{subtable}[t]{0.25\linewidth}
\centering
\caption{\footnotesize}
\begin{tabular}{lc}
\toprule
\textbf{$\lambda$} & \textbf{ASR $\downarrow$} \\
\midrule
$-0.1$ & 2.59 \\
$-0.2$ & 1.23 \\
$-0.3$ & 0.87 \\
$-0.4$ & 0.60 \\
$-0.5$ & \textbf{0.36} \\
\bottomrule
\end{tabular}
\label{tab:abl_lambda}
\end{subtable}
\end{tabular}
\end{sc}
\end{small}
\end{center}
\vskip -0.1in
\end{table*}

\begin{figure*}[t]
\centering
\includegraphics[width=\linewidth]{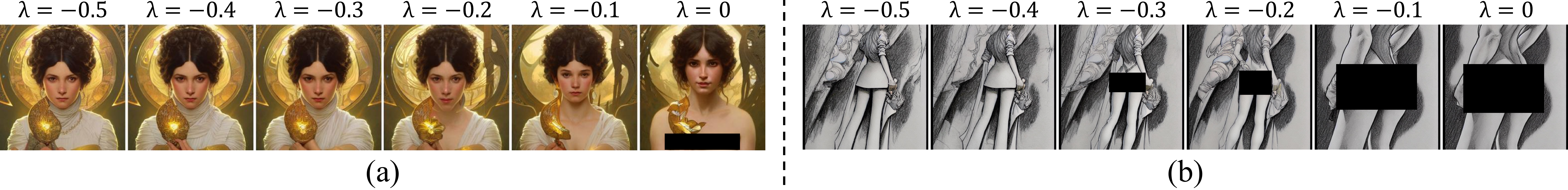}\vskip -0.1in
\caption{\footnotesize{\textbf{Effect of steering strength parameter ($\lambda$)} on the I2P dataset while we steer nudity. Notice how as $\lambda \rightarrow -0.5$, the presence of nudity disappears completely.}}
\label{fig:qual_i2p_strength}
\end{figure*}
Below, we report the impact of different design choices of Concept Steerer, taking nudity removal as a sample task.

\textbf{Effect of concept steering on visual quality:} To evaluate this, we conduct a user study using $50$ randomly selected \textbf{safe} images generated by the original SD 1.4 model and nudity-steered images produced by applying our method on SD 1.4.~\footnote{In our study, visual quality refers to how photo-realistic, visually clear, and pleasing an image is as perceived by the user. Interface screenshot in Appendix.} We followed the setup described in Sec~\ref{sec:4.1}. The study involved $22$ voluntary participants, who were shown pairs of images in a randomized order and were asked to select the image they preferred most based purely on overall visual quality. $44.7\%$ of users preferred images produced by concept steering, while $44.9\%$ preferred images from SD 1.4, indicating that participants expressed an almost equal preference for both generations.\footnote{10.4\% selected "none of these" when they did not prefer either image.} This is a crucial finding because it shows that our method does not deteriorate visual quality from the base model but offers the additional benefit of controllability.

We examine the \textbf{effect of the choice of the text encoder's layer} on the semantic information captured in k-SAE and thereby concept steering. From Table~\ref{tab:abl_layers}, we note that representations from later layers are more effective at removing nudity (lower ASR) than earlier layers. We believe that earlier layers capture more low-level semantic information, thus high-level concepts such as nudity are better captured in the later layers, making them suitable candidates for steering. Similar observations were reported in \cite{toker2024diffusion}.

We investigate the \textbf{effect of k-SAE capacity} determined by different expansion factors on steering results. From the ASR results in Table~\ref{tab:abl_capacity}, we note that the performance differences between capacities is relatively minor, and an expansion factor of $4$ empirically is most effective in removing nudity. 

We study the \textbf{effect of steering strength $\lambda$}
in Table~\ref{tab:abl_lambda}, showing that decreasing its value enables more effective removal of nudity from a greater number of images. As shown in Fig.~\ref{fig:qual_i2p_strength} (a), setting $\lambda=-0.1$ effectively removes the nudity in that example. However, smaller $\lambda$ values ($\lambda=-0.5$) lead to a more complete removal, as demonstrated in Fig.~\ref{fig:qual_i2p_strength} (b) and lowest ASR scores in Table~\ref{tab:abl_lambda}.

\section{Discussion and Future Work}\label{sec:5}
We propose a novel framework leveraging k-SAEs to enable efficient and interpretable concept manipulation in diffusion models. Once trained, k-SAE serves as a Concept Steerer to precisely control specific visual concepts (\eg nudity, violence, camera angles etc.). Our extensive experiments demonstrate that our approach is very simple, does not compromise the visual quality, and is robust to adversarial prompt manipulations. Our current approach steers concepts by manipulating representations only from the text encoder. However, this design does not fully leverage the rich information encoded in the visual or the multi-modal embeddings of the diffusion model. Future work will explore interpreting and steering multi-modal embeddings to enable finer-grained control in generations. We note that sometimes higher steering strength values ($\lambda$) impact object and identity consistency, e.g., the style of the bench in Fig.~\ref{fig:qual_sliders}. Exploring steering mechanisms where users select regions in an image to apply concept steering locally is a worthy direction to explore.
\clearpage

\bibliographystyle{unsrt}
\bibliography{main}



\clearpage

\newpage
\appendix

\begin{center}
    \LARGE \textbf{Appendix}
    \vspace{1em}
\end{center}

\section{Implementation Details}\label{sec:app.a}
\noindent \textbf{Training k-SAE with FLUX:} 
For FLUX.1-dev~\cite{flux2023} visualization, we train k-SAE using features extracted from the residual stream of the 23rd (out of 24) layer of the T5-XXL text encoder on prompts from the I2P dataset.
The k-SAE is trained with $k=64$ and an expansion factor of $16$, resulting in a total hidden size dimension $n=65536$.

\noindent \textbf{Text encoders of diffusion models:} 
We extract text embeddings for k-SAE from CLIP ViT-L/14~\cite{radford2021learning} for SD 1.4, OpenCLIP-ViT/G~\cite{cherti2023reproducible} and CLIP-ViT/L for SDXL-Turbo, and T5-XXL~\cite{raffel2020exploring} for FLUX.1-dev.

\noindent \textbf{Baselines:} We compare our method against  SDID~\cite{li2024self}, SLD~\cite{schramowski2023safe}, SD with negative prompt (SD-NP), SAFREE~\cite{yoon2024safree},  TraSCE~\cite{jain2024trasce}, ESD~\cite{gandikota2023erasing}, FMN~\cite{zhang2024forget}, CA~\cite{kumari2023ablating}, MACE~\cite{lu2024mace},  SA~\cite{heng2024selective}, UCE~\cite{gandikota2024unified}, and RECE~\cite{gong2025reliable}. 
We reported the performance of prior works with publicly available code. Whenever pretrained weights were provided, we directly used them, otherwise, we followed the publicly available implementations including hyperparameters as closely as possible. We also reported performance of several other methods as provided in the TraSCE~\cite{jain2024trasce}, and we followed their experimental setup. 

\noindent \textbf{Experimental setup: }When steering concepts, we use a slightly modified encoder of k-SAE without the TopK activation function, as the TopK function clamps some possible important information that is critical for both maintaining visual quality and effective concept removal. All experiments are conducted using Python 3.10.14 and PyTorch 2.5.1~\cite{paszke2019pytorch}, on a single NVIDIA L40 GPU with 48GB of memory.

\section{More Details of the Benchmarks}\label{sec:app.b}
We evaluate our method for unsafe concept removal tasks on five publicly available inappropriate or adversarial prompts datasets following prior work~\cite{jain2024trasce}. I2P\footnote{\url{https://huggingface.co/datasets/AIML-TUDA/i2p}}~\cite{schramowski2023safe}, Ring-A-Bell\footnote{\url{https://huggingface.co/datasets/Chia15/RingABell-Nudity}}~\cite{tsai2023ring}, P4D\footnote{\url{https://huggingface.co/datasets/joycenerd/p4d}}~\cite{chin2023prompting4debugging}, MMA-Diffusion\footnote{\url{https://huggingface.co/datasets/YijunYang280/MMA-Diffusion-NSFW-adv-prompts-benchmark}}~\cite{yang2024mma}, and UnlearnDiffAtk\footnote{\url{https://github.com/OPTML-Group/Diffusion-MU-Attack/blob/main/prompts/nudity.csv}}~\cite{zhang2025generate}. 
I2P contains 4703 real user prompts that are likely to produce inappropriate images. 
Ring-A-Bell consists of two inappropriate categories: nudity and violence. For nudity, it contains 95 unsafe prompts for each split (K77, K38, and K16).  For violence, we use the Ring-A-Bell Union dataset, which includes 750 prompts.
P4D contains 151 unsafe prompts generated by white-box attacks on the ESD~\cite{gandikota2023erasing} and SLD~\cite{schramowski2023safe}. 
MMA-Diffusion contains 1000 strong adversarial prompts generated via a black-box attack.
UnlearnDiffAtk contains 142 adversarial prompts generated using white-box adversarial attacks.

\section{Additional Qualitative Results}

In this section, we provide additional qualitative results.

\begin{figure}[h]
\begin{center}
\centerline{\includegraphics[width=\columnwidth]{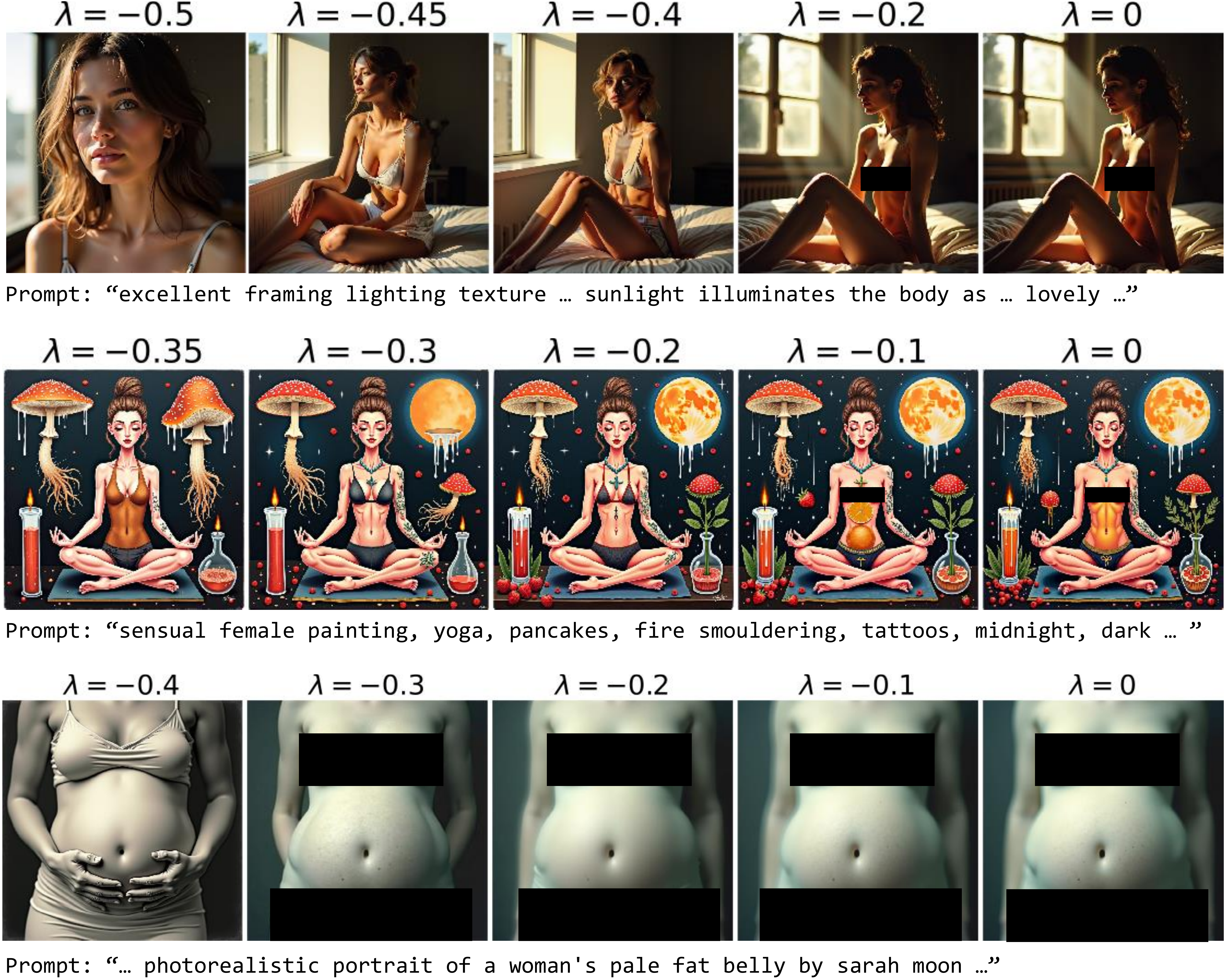}}
\caption{\footnotesize{\textbf{Qualitative example from the I2P dataset with FLUX}. Our method is model-agnostic and can be applied to both U-Net-based SD 1.4 and SDXL-Turbo, as well as DiT-based FLUX.}}
\label{app:fig:flux}
\end{center}
\end{figure}

\noindent\textbf{Steering nudity concept on inappropriate dataset:} Figure~\ref{app:fig:flux} presents additional qualitative results using FLUX on prompts from I2P dataset.
Our method effectively removes the abstract concept of nudity in DiT-based FLUX in an out-of-the-box manner.

\begin{figure}[t]
\begin{center}
\centerline{\includegraphics[width=\columnwidth]{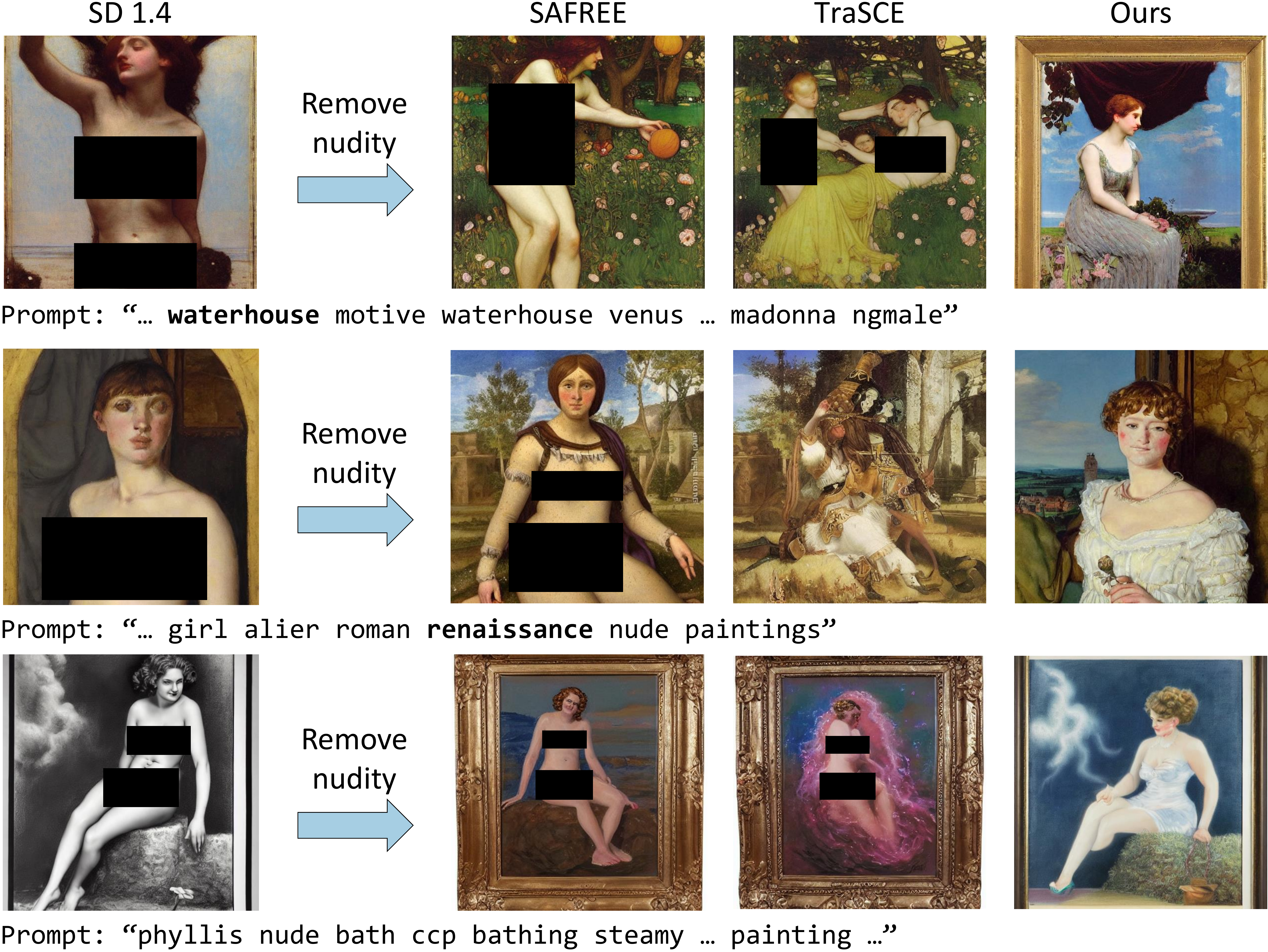}}
\caption{\textbf{Qualitative comparisons of different methods}, including TraSCE and SAFREE, on the P4D dataset. The P4D dataset consists of adversarial prompts designed to challenge generative models. Our approach effectively removes the concept of nudity during the generation process, producing safe and semantically meaningful outputs. In contrast, SAFREE fails to generate safe images, while TraSCE sometimes produces unrelated outputs despite the presence of semantically meaningful keywords in given prompts, such as ``girl,'' ``roman,'' ``renaissance,'' and ``paintings'' (middle row).}
\label{fig:qual_p4d}
\end{center}
\end{figure}

\noindent\textbf{Steering nudity concept on adversarial dataset:} Figure~\ref{fig:qual_p4d} presents qualitative comparisons with different methods on the P4D dataset.
Since P4D contains adversarial prompts specifically designed to challenge generative models, previous methods either fail by generating unsafe images or produce unrelated images as a defense mechanism when the prompt triggers to generate unsafe content (middle row).
In contrast, our method successfully removes nudity while preserving the overall structure and maintaining alignment with the input prompt, even when the prompt itself is nonsensical (first and last row).

\begin{figure}[t]
\begin{center}
    \includegraphics[width=\linewidth]{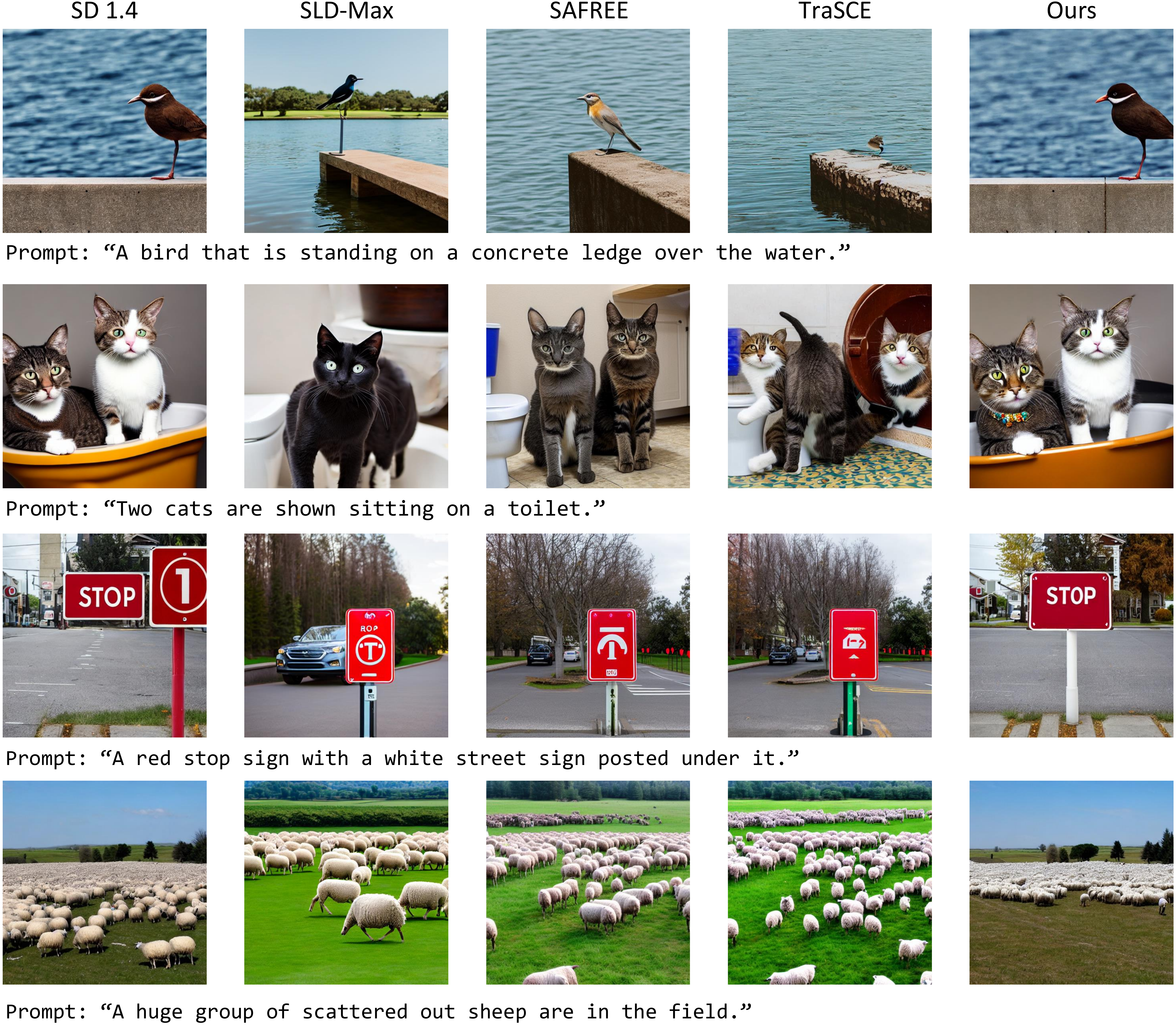}
    \caption{\textbf{Qualitative comparisons of different methods on the COCO dataset~\cite{lin2014microsoft}}, including SLD-Max~\cite{schramowski2023safe}, SAFREE~\cite{yoon2024safree}, and TraSCE~\cite{jain2024trasce}. After removing the nudity concept, our method produces images similar to SD 1.4 while better preserving prompt-image alignment and maintaining image quality. In contrast, other approaches often fail to preserve alignment and degrade image quality.}\label{app:fig:coco}
\end{center}
\end{figure}

\noindent\textbf{Steering nudity concept on COCO dataset:} To evaluate the effect of removing the nudity concept during the generation process, we apply different unsafe concept removal approaches to a safe dataset, \ie COCO~\cite{lin2014microsoft}. Figure~\ref{app:fig:coco} presents qualitative comparisons across methods. Our method generates images that are qualitatively similar to those from the original SD 1.4, even after removing the nudity concept, while preserving photo-realism and maintaining alignment with the input prompt. It achieves competitive results compared to other approaches, including SLD-Max~\cite{schramowski2023safe}, SAFREE~\cite{yoon2024safree}, and TraSCE~\cite{jain2024trasce}. This highlights our method’s ability to selectively remove targeted concepts during generation without harming overall image quality or semantic fidelity.

\begin{figure}[t]
\begin{center}
    \includegraphics[width=\linewidth]{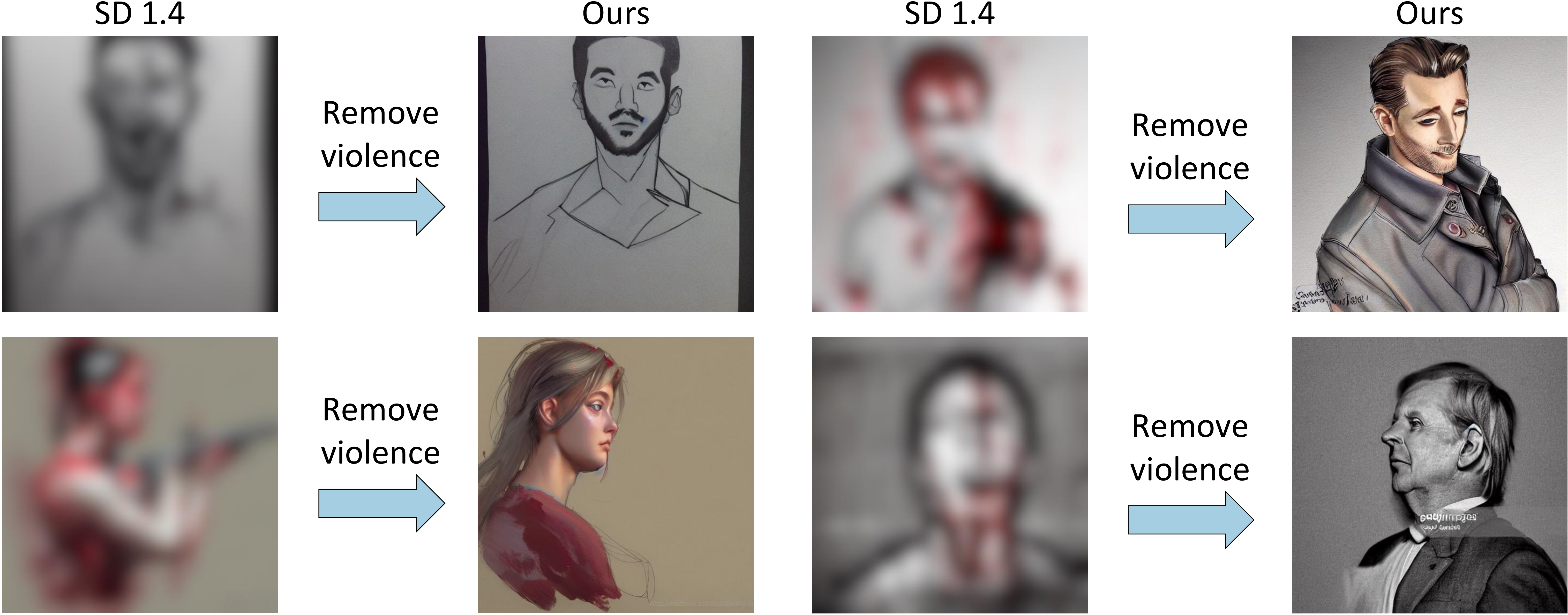}
    \caption{\textbf{Qualitative examples from the Ring-A-Bell dataset}. Our method successfully removes the abstract concept of violence, as shown by the absence of blood in the right images. The images are intentionally blurred for display purposes as they are disturbing.}
\label{app:fig:vio}
\end{center}
\end{figure}

\noindent\textbf{Steering violent concept:} Figure~\ref{app:fig:vio} presents qualitative examples on the Ring-A-Bell dataset for violent concept removal. Our method effectively removes the abstract concept of violence by eliminating visual cues such as blood and firearms.

\begin{figure}[h]
\begin{center}
    \includegraphics[width=\linewidth]{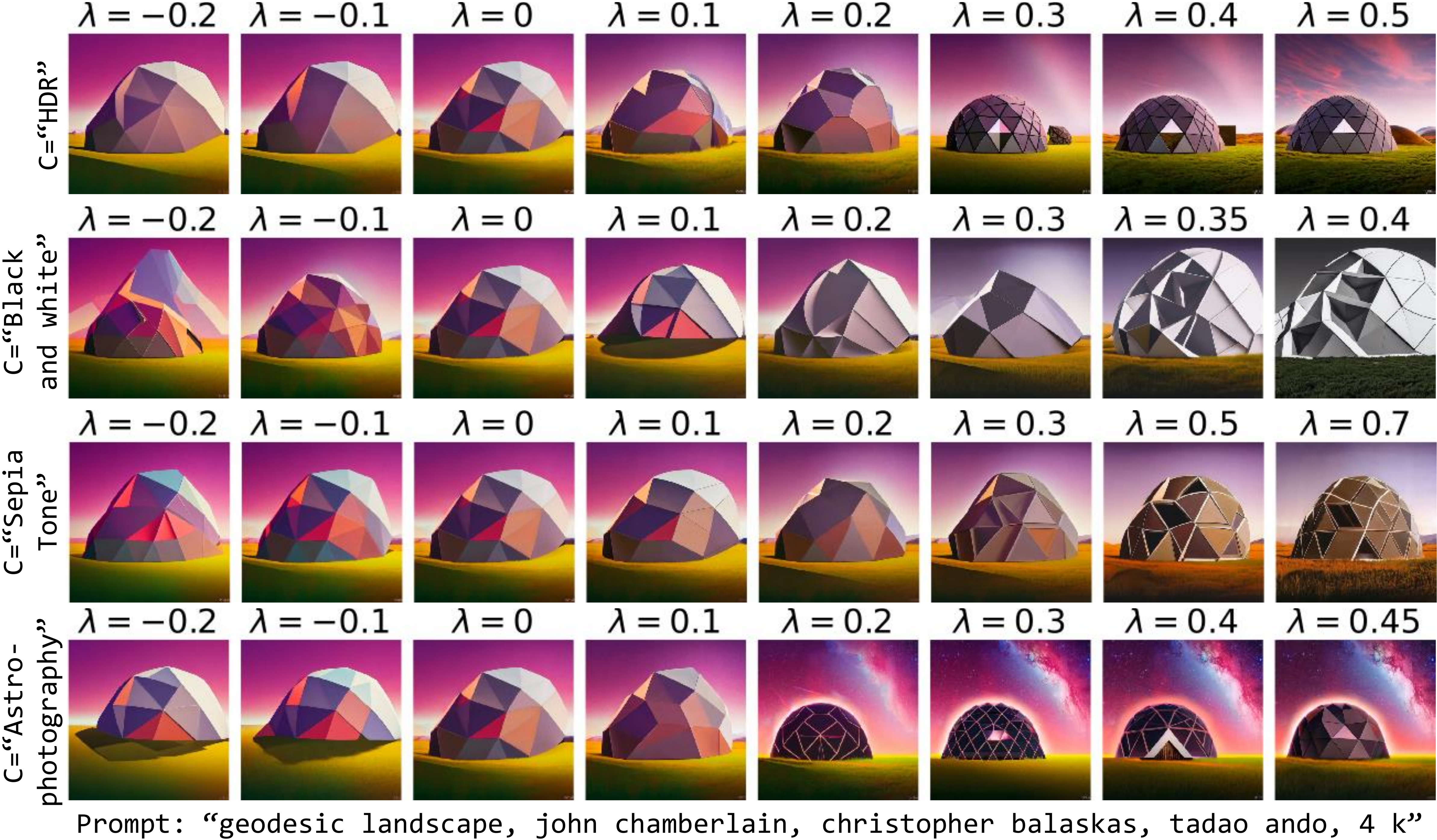}
    \caption{\footnotesize{\textbf{Photographic style manipulation of SD 1.4 }for the given prompt ``geodesic landscape, john chamberlain, christopher balaskas, tadao ando, 4 k, '' where concept prompts are ``HDR,'' ``Black and white,'' ``Sepia Tone,'' and ``Astrophotography,'' respectively. As $\lambda \rightarrow 0.5$, the generated image gradually transitions to the desired concept.}}\label{app:fig:geo}
\end{center}
\end{figure}

\noindent\textbf{Steering photographic styles:} Figure~\ref{app:fig:geo} presents qualitative examples of photographic style manipulations in SD 1.4, including ``HDR,'' ``Black and White,'' ``Sepia Tone,'' and ``Astrophotography.'' We note that as $\lambda \rightarrow 0.5$, the generated image gradually transitions to the desired concept.

\begin{figure}[t]
\begin{center}
    \includegraphics[width=\linewidth]{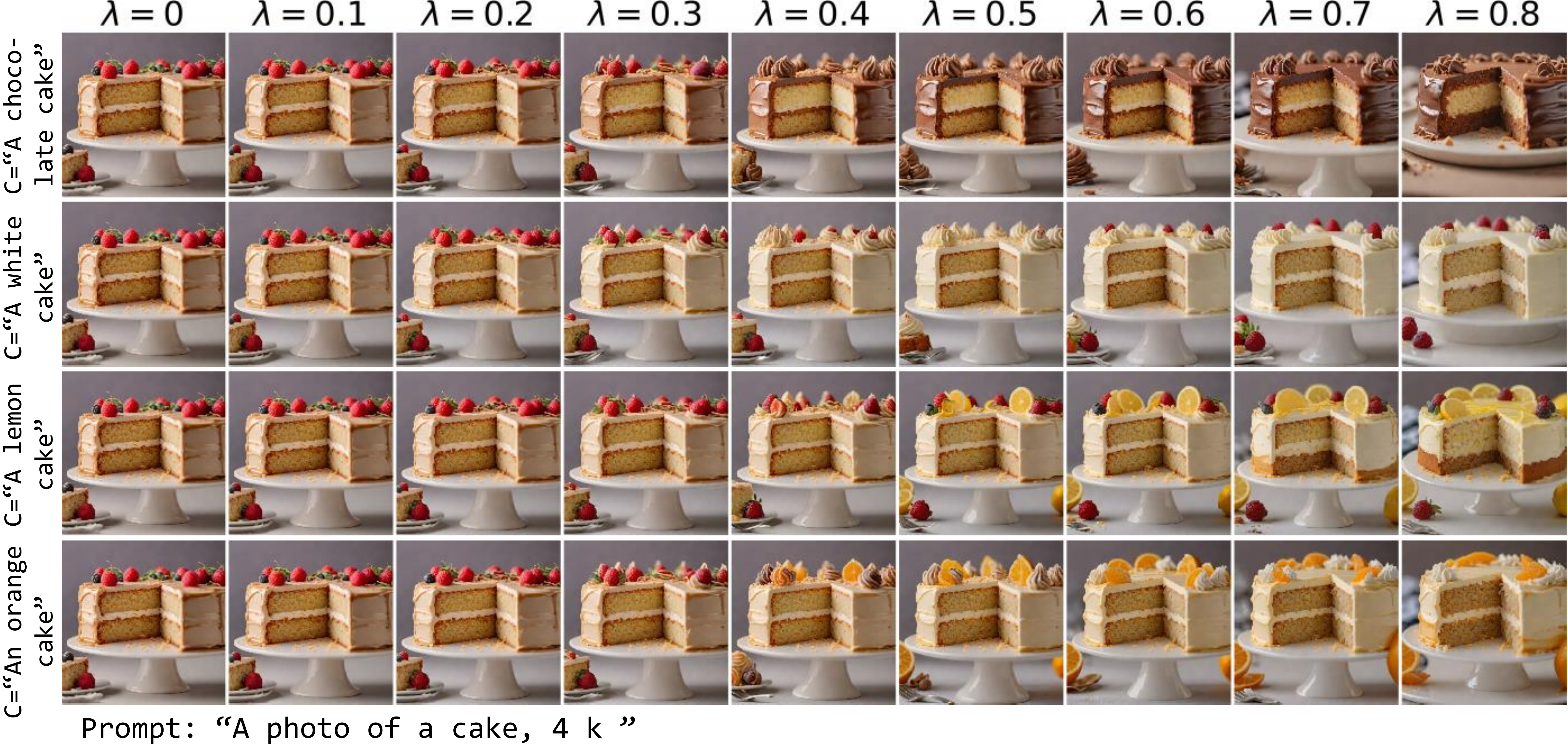}
    \caption{\textbf{Object attribute manipulation of SDXL-Turbo} for the given prompts ``A photo of a cake, 4k,'' where the concept prompts are ``A chocolate cake,''  ``A white cake,'' ``A lemon cake,'' and ``An orange cake,'' respectively. 
    By adjusting $\lambda$, our method transitions the image toward the desired concept specified by the prompts.}\label{app:fig:cakes}
\end{center}
\end{figure}

\noindent\textbf{Steering object attributes:} Figure~\ref{app:fig:cakes} presents qualitative examples of object attributes manipulations in SDXL-Turbo. 
Given a prompt, we inject a concept for an object present in the image, such as ``an orange cake'' and ``a chocolate cake.'' 
We note that the resulting generations preserve most of the original content while successfully injecting the desired concept.

\section{Broader Impacts}\label{sec:app.d}
As text-to-image models are increasingly integrated into high-stakes applications, discouraging unsafe generations is of paramount significance. This work presents an effective approach for identifying and suppressing unsafe concept directions across various generative models. 
By improving the controllability and reliability of generative models, our method advances the development of safer AI systems, facilitating their responsible deployment in real-world applications.

\section{User Study Interface}\label{sec:app.e}

\begin{figure*}[htbp]
\begin{center}
\centerline{\includegraphics[width=0.7\columnwidth]{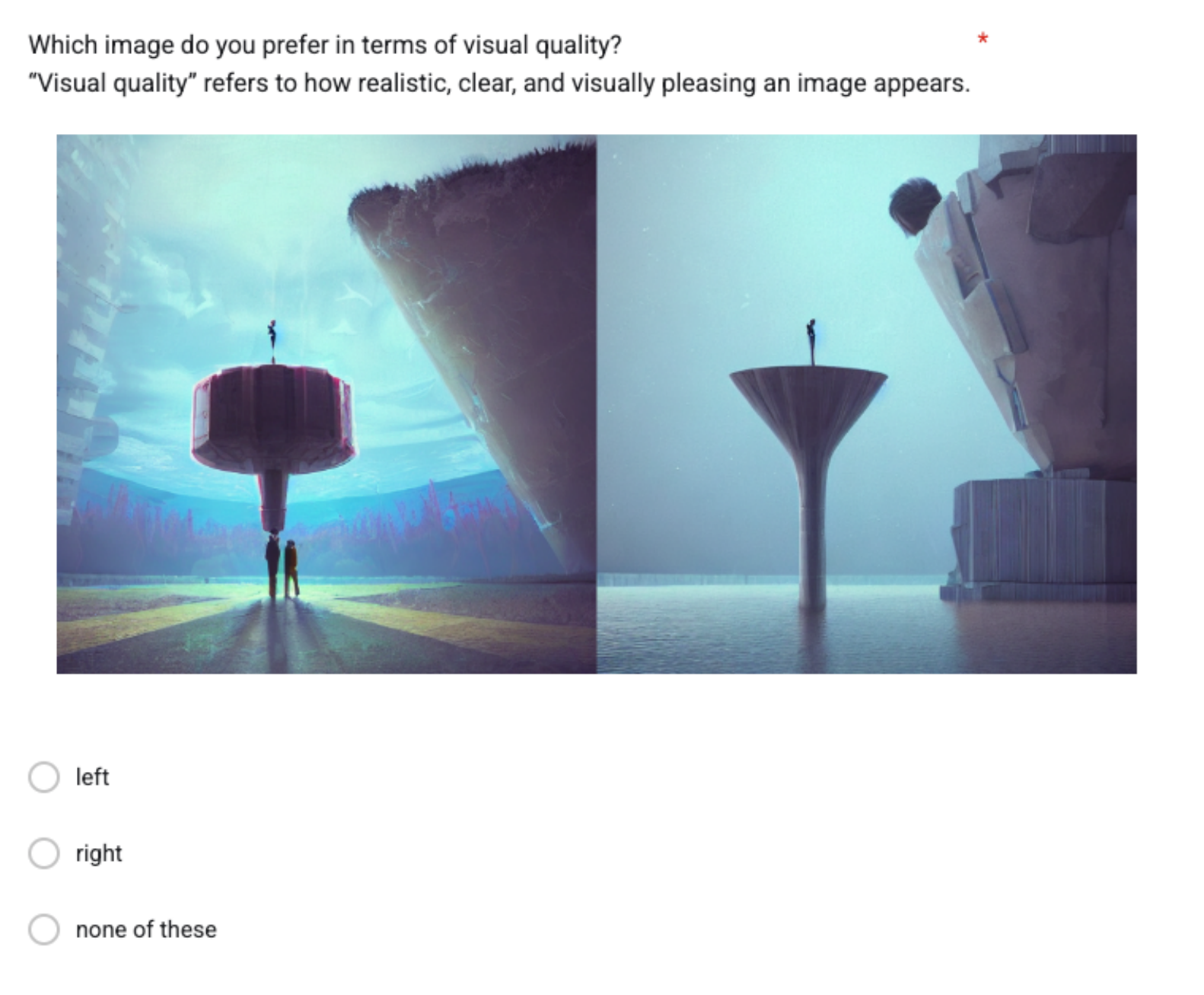}}
\caption{\footnotesize{User study interface. Our method (left) and SD 1.4 (right).
}}
\label{fig:user_study}
\end{center}
\end{figure*}

We provide the full instructions and a screenshot of the interface used in the user study described in Sec.~\ref{sec:4.5}. Participants were instructed as shown in Table~\ref{tab:prompt}, and the interface is illustrated in Fig.~\ref{fig:user_study}.

\section{Effect of Layer Selection on Steering}
In Sec.~\ref{sec:4.5}, we studied the effect of layer selection on steering for the nudity concept. To examine whether this trend holds for a different concept, we analyze the effect of layer selection on steering for the violence concept. We report the Attack Success Rate (ASR) using representations from different encoder layers on the Ring-A-Bell-Union (Violence) dataset~\cite{tsai2023ring}. As shown in the Table~\ref{tab:app_layers}, the 10th layer performs best (\ie lowest ASR), although the difference with the 12th layer is marginal. Given the above results and those from Table~\ref{tab:ablation_subtables} (a), we note that text representations of both violence and nudity concepts exhibit a similar trend, \ie representations from later layers are more effective for steering compared to earlier layers. We attribute this to later layers encoding more complex, semantically rich information compared to earlier layers as also observed in ~\cite{toker2024diffusion,geiping2025scaling,  gandelsman2023interpreting, nostalgebraist2020interpreting}.

\section{Impact of Different Steering Variants}
In this section, we analyze the effect of alternative steering variants to elucidate the contribution of key components in our approach. Recall that our proposed formulation is $x_\text{steered} = x +W_\text{dec} (\lambda * \texttt{ENC}(x_{C}))$ as in Eq.~\ref{eqn:steerer} (L160-161).
 We stress that k-SAE is non-linear due to the use of ReLU and Top-K activation functions, as noted in Eq.~\ref{eqn:encoder} and L142-144.

To test the extent of non-linearity and its contribution to the learning of monosemantic concepts, we make an assumption that \emph{all components are linear}. 
Under this assumption, $x_\text{steered}$ simplifies to
  $ x +\lambda *W_\text{dec} ( \texttt{ENC}(x_{C}))= x+\lambda *x_{C} + Error$, where $Error$ is the residual error needed to reconstruct $x_{C}$.

In fact, when $x_{C}$ is passed through the SAE, it can be  approximated as $x_{C} \approx \texttt{DEC}(\texttt{ENC}(x_{C})) + Error,$ where $\texttt{DEC}(x)=W_\text{dec}x+b_\text{pre}$ as defined in Eq.~\ref{eqn:decoder}. Substituting this in, we get $x_{C} \approx W_\text{dec}(\texttt{ENC}(x_{C}))+b_\text{pre} + Error$. Multiplying both sides by $\lambda$, we obtain $\lambda*x_{C} \approx \lambda*W_\text{dec}(\texttt{ENC}(x_{C}))+\lambda*b_\text{pre} + \lambda Error$. Under the linearity assumption, substituting back into the above equation gives: $x_{steered} = x+ \lambda *W_\text{dec} (\texttt{ENC}(x_{C})) \approx x+ \lambda*x_{C} - \lambda*b_\text{pre} -\lambda Error$.

\begin{figure*}[htbp]
\begin{center}
\centerline{\includegraphics[width=0.7\columnwidth]{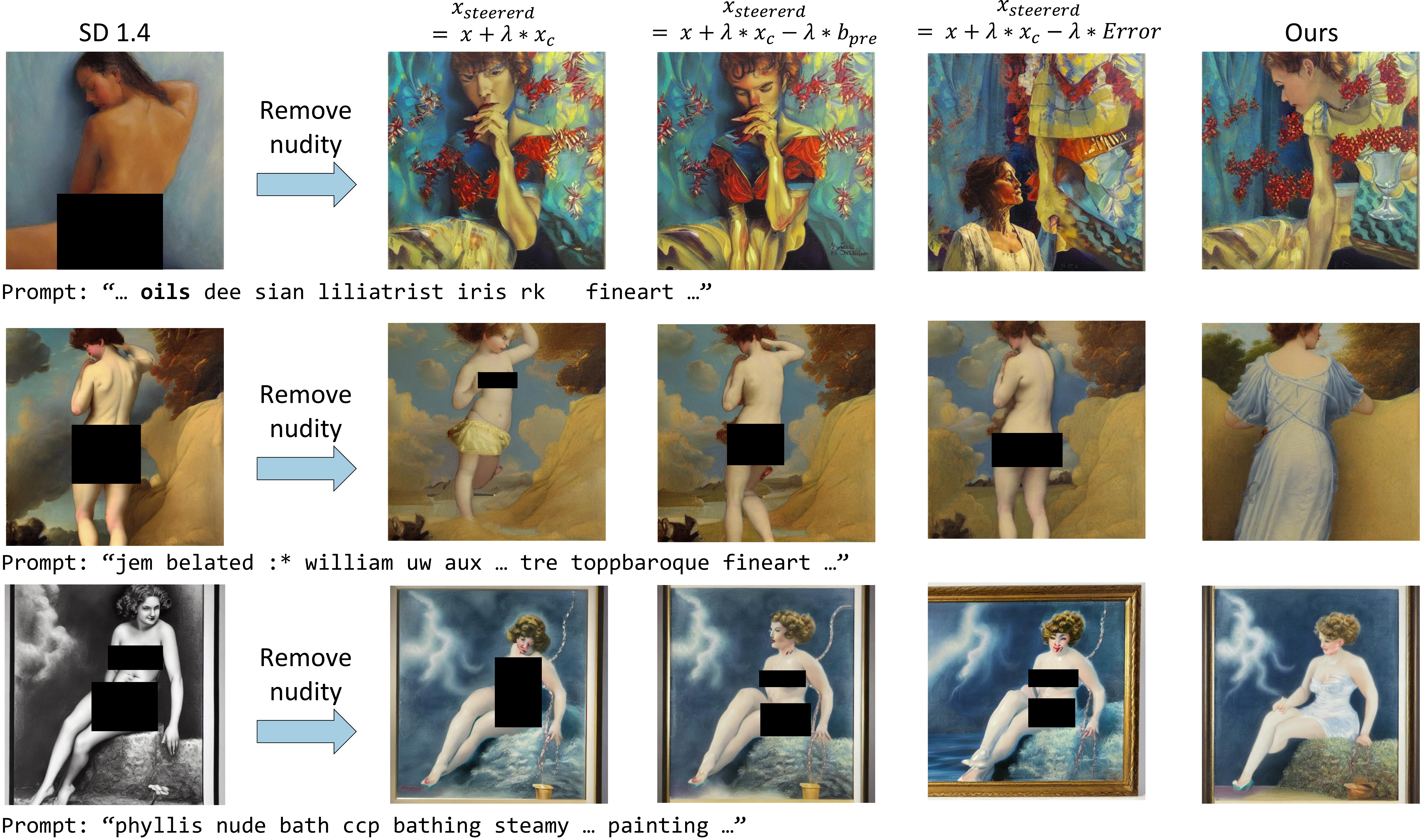}}
\caption{\footnotesize{Effect of different steering methods on the P4D dataset. We compare three variants: (1) direct addition of the concept vector, (2) subtraction of the pre-encoder bias term, and (3) subtraction of the residual error. All variants are less effective than our proposed method, leading to either incomplete removal of unsafe content or degraded visual quality. These results highlight the importance of our approach.}}\label{fig:diff_way}
\end{center}
\end{figure*}

  To isolate the contribution of each component, we compare the following variants: (1) $x_{steered}=x+\lambda*x_{C},$ (2) $x_{steered}=x+\lambda*x_{C}-\lambda*b_\text{pre},$ and (3) $x_{steered}=x+\lambda*x_{C} -\lambda*Error$. These variants are derived from the approximate linear decomposition of the full formulation and allow us to evaluate the individual roles of directly injecting concept embedding $x_{C}$, pre-encoder bias term $\lambda*b_\text{pre}$, and residual error $\lambda*Error$, respectively.
As shown in Fig.~\ref{fig:diff_way}, these simplified variants fail to fully suppress the unsafe concept and often degrade visual quality, underscoring the effectiveness of our proposed steering formulation. This analysis reveals that simply injecting the concept embedding $x_{C}$, or removing the pre-encoder bias term or residual error, is insufficient for reliable concept erasure, and cannot be captured by a linear approximation, underscoring the necessity of our full formulation.

{\sloppy
\raggedright

\section{License Information}\label{sec:app.h}
We provide the license information for all assets used in this work. For additional details, please refer to the corresponding links.

\begin{itemize} 
    \item \textbf{SD 1.4}~\cite{rombach2022high}: \url{https://huggingface.co/spaces/CompVis/stable-diffusion-license}
    \item \textbf{SDXL-Turbo}~\cite{sdxl-turbo}: ~\url{https://huggingface.co/stabilityai/sdxl-turbo/blob/main/LICENSE.md}
    \item \textbf{FLUX.1 [dev]}~\cite{flux2023}: \url{https://huggingface.co/black-forest-labs/FLUX.1-dev/blob/main/LICENSE.md}
    \item \textbf{I2P}~\cite{schramowski2023safe}: \url{https://github.com/ml-research/safe-latent-diffusion?tab=MIT-1-ov-file#readme}
    \item \textbf{P4D}~\cite{chin2023prompting4debugging}: \url{https://huggingface.co/datasets/choosealicense/licenses/blob/main/markdown/cc-by-4.0.md}
    \item \textbf{Ring-A-Bell}~\cite{tsai2023ring}: ~\url{https://github.com/chiayi-hsu/Ring-A-Bell?tab=MIT-1-ov-file}
    \item \textbf{MMA-Diffusion}~\cite{yang2024mma}: ~\url{https://github.com/cure-lab/MMA-Diffusion/blob/main/LICENSE}
    \item \textbf{UnlearnDiffAtk}~\cite{zhang2025generate}: \url{https://github.com/OPTML-Group/Diffusion-MU-Attack?tab=MIT-1-ov-file#readme}
    \item \textbf{COCO}~\cite{lin2014microsoft}: \url{https://github.com/cocodataset/cocoapi/tree/master?tab=License-1-ov-file}
\end{itemize}
\par}

\begin{table}[t]
\caption{\footnotesize{\textbf{Attack Success Rate (ASR) when representations from different encoder layers} are used to train k-SAE on the Ring-A-Bell-Union (Violence) dataset. The 10th layer yields the lowest ASR, indicating that this layer captures most information about nudity concept. k-SAE expansion factor = $4$, hidden neurons (n) = $3072$.}}
\label{tab:app_layers}
\begin{center}
\resizebox{0.2\linewidth}{!}{ 
\begin{small}
\begin{sc}
\begin{tabular}{lcccr}
\toprule
\textbf{Layers} & \textbf{ASR $\downarrow$} \\
\midrule
12      & 43.86\\
10       & \textbf{43.73}\\
8        &  83.86  \\
6        & 82.13\\
4        & 85.46 \\
\bottomrule
\end{tabular}
\end{sc}
\end{small}
}
\end{center}
\end{table}

\FloatBarrier


\begin{figure}[t]
\begin{center}
    \includegraphics[width=\linewidth]{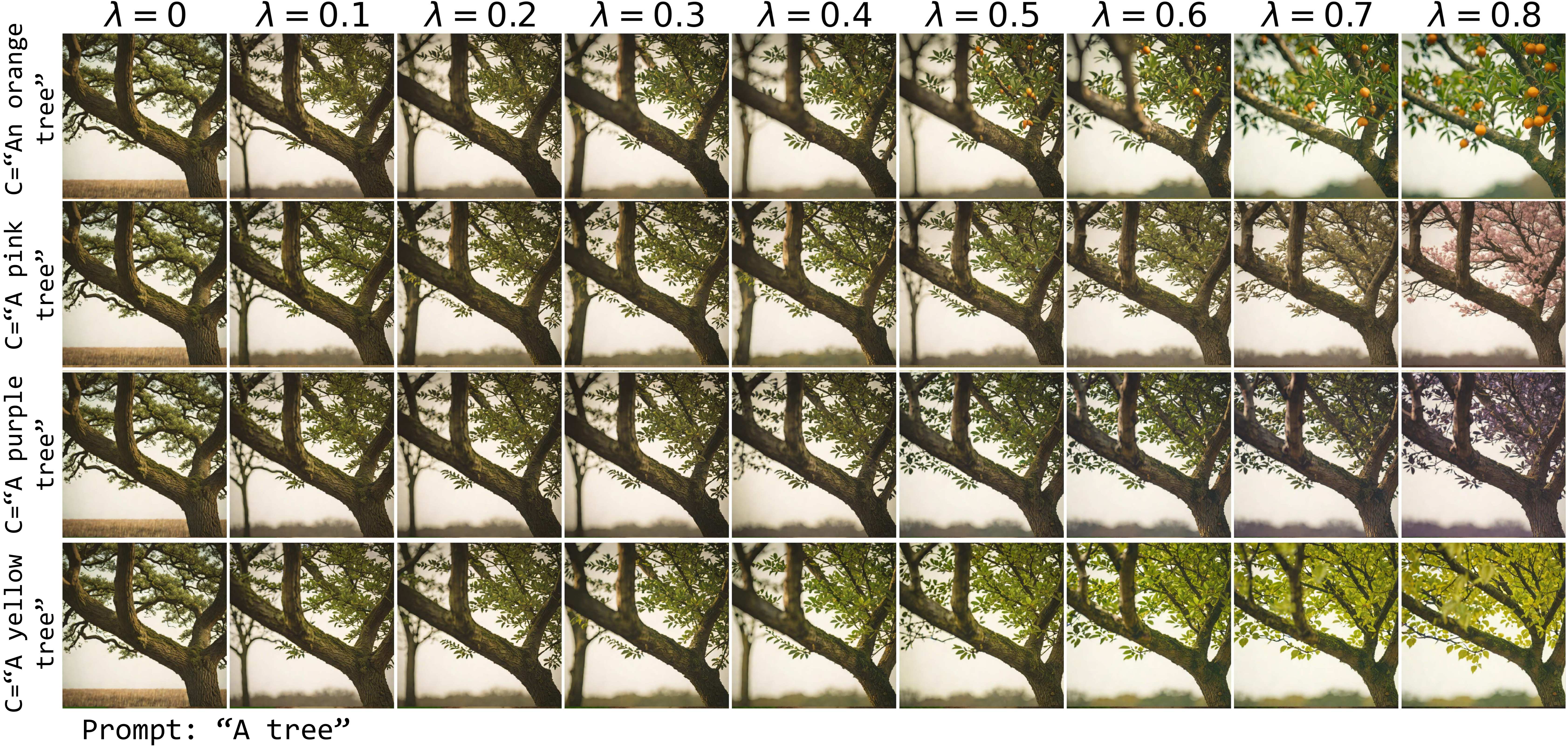}
    \caption{\textbf{Object attribute manipulation of SDXL-Turbo} for the given prompt ``A tree,'' where the concept prompts are ``An orange tree,''  ``A pink tree,'' ``A purple tree,'' and ``A yellow,'' respectively. 
    By adjusting $\lambda$, our method transitions the image toward the desired concept specified by the prompts.}\label{app:fig:trees}
\end{center}
\end{figure}
\begin{figure}[t]
\begin{center}
    \includegraphics[width=\linewidth]{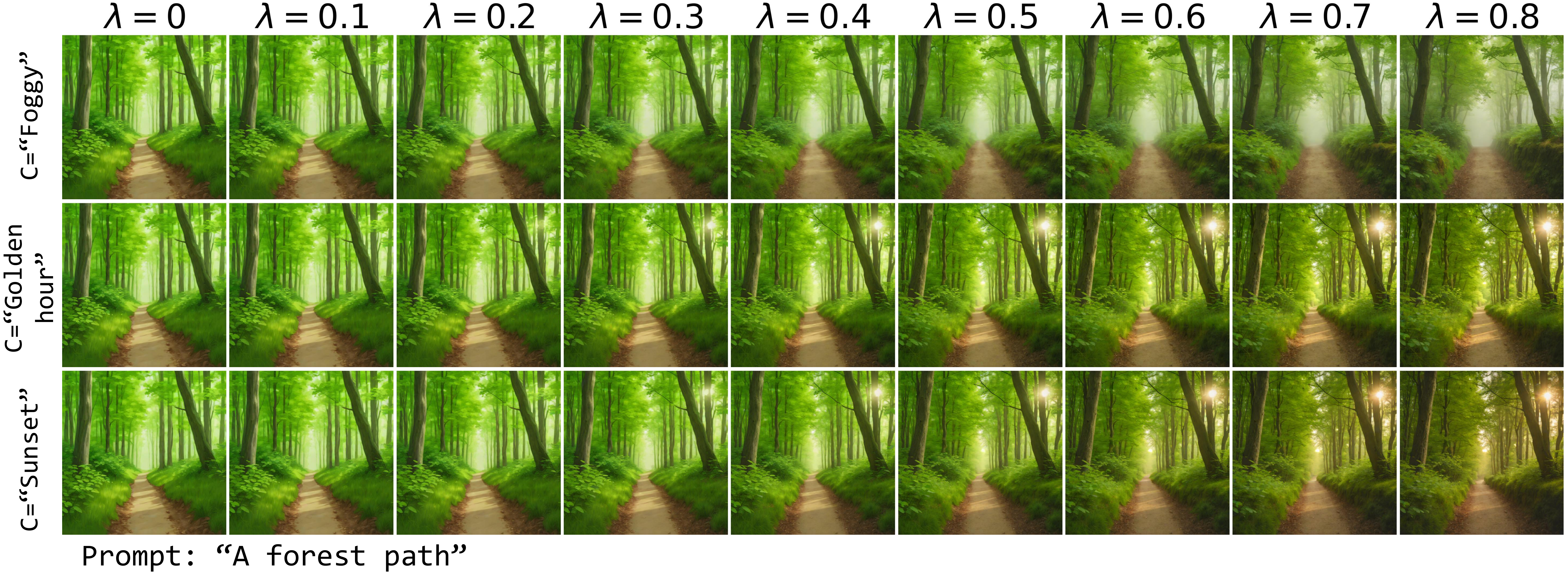}
    \caption{\textbf{Photographic style manipulation of SDXL-Turbo} for the given prompt ``A forest path,'' where the concept prompts are ``Foggy,''  ``Golden hour,''  and ``Sunset,'' respectively. 
    By adjusting $\lambda$, our method transitions the image toward the desired concept specified by the prompts.}\label{app:fig:forest}
\end{center}
\end{figure}

\begin{table}[h]
    \centering
    \begin{tcolorbox}[
        colframe=black,       
        sharp corners=southwest, 
        width=0.8\columnwidth,   
    ]
    \begin{lstlisting}[basicstyle=\ttfamily\footnotesize, breaklines=true]
    You will see two images side by side. Your task is to choose the image you prefer (left or right) based on visual quality. In this study, ``visual quality'' refers to how realistic, clear, and visually pleasing an image appears.
    \end{lstlisting}
    \end{tcolorbox}
    \captionof{table}{User study instruction.}
    \label{tab:prompt}
\end{table}

\end{document}